\documentclass{article}

\usepackage[preprint]{neurips_2026}

\usepackage[utf8]{inputenc} % allow utf-8 input
\usepackage[T1]{fontenc}    % use 8-bit T1 fonts
\usepackage{hyperref}       % hyperlinks
\usepackage{url}            % simple URL typesetting
\usepackage{booktabs}       % professional-quality tables
\usepackage{amsfonts}       % blackboard math symbols
\usepackage{nicefrac}       % compact symbols for 1/2, etc.
\usepackage{microtype}      % microtypography
\usepackage[table]{xcolor}         % colors
\usepackage{graphicx}
\usepackage{tabularray}
\usepackage{caption}
\usepackage{subcaption}
\usepackage[utf8]{inputenc}
\DeclareUnicodeCharacter{2212}{\ensuremath{-}}

\newcommand{\rA}[1]{\cellcolor{green!35}\textbf{#1}}
\newcommand{\rB}[1]{\cellcolor{green!18}#1}
\newcommand{\rC}[1]{\cellcolor{green!8}#1}
\newcommand{\rD}[1]{#1}
\newcommand{\rE}[1]{\cellcolor{red!8}#1}
\newcommand{\rF}[1]{\cellcolor{red!18}#1}
\newcommand{\rG}[1]{\cellcolor{red!35}#1}

\title{EarthShift: a benchmark for measuring robustness to real-world distribution shifts in Earth observation}

\author{%
  Kelsey Doerksen \\
  School of Computing and Augmented Intelligence
  \\
  Arizona State University\\
  \texttt{kdoerkse@asu.edu} \\
  \And
    Hannah Kerner \\
    School of Computing and Augmented Intelligence \\
    Arizona State University \\
}

\begin{document}

\maketitle

\begin{abstract}
Current Earth observation benchmarks focus on measuring performance on diverse tasks and applications, typically measuring generalization in-distribution. But when models are deployed, they must generalize to myriad out-of-distribution scenarios, such as new time periods, geographies, scales, and sensors. We introduce EarthShift: the first public testbed for benchmarking robustness across multiple realistic distribution shifts encountered in remote sensing. EarthShift enables users to measure distributional robustness by comparing performance in- and out-of-distribution using datasets from paired datasets from different sources, temporal windows, geographic locations, and sensors. Our experiments on 8 geospatial foundation models (GFMs) and 11 tasks covering 5 shift types show that GFMs consistently perform 15-20\% worse out-of-distribution on average regardless of model architecture, size, pre-training or fine-tuning strategy. We show that GFM robustness is similar to that of generic vision foundation models, and even fully-supervised models. This highlights a need for future research to strive for improvements in distributional robustness, not just performance, which can be benchmarked using EarthShift. We release our code and datasets to provide a testbed to guide future work to create foundation models that are robust and reliable in real-world applications. Code and data for EarthShift are available at: \href{https://earthshift.github.io}{earthshift.github.io}.
\end{abstract}

\section{Introduction}
\label{sec:intro}

\begin{figure*}[t]
    \centering
    \includegraphics[width=0.8\textwidth]{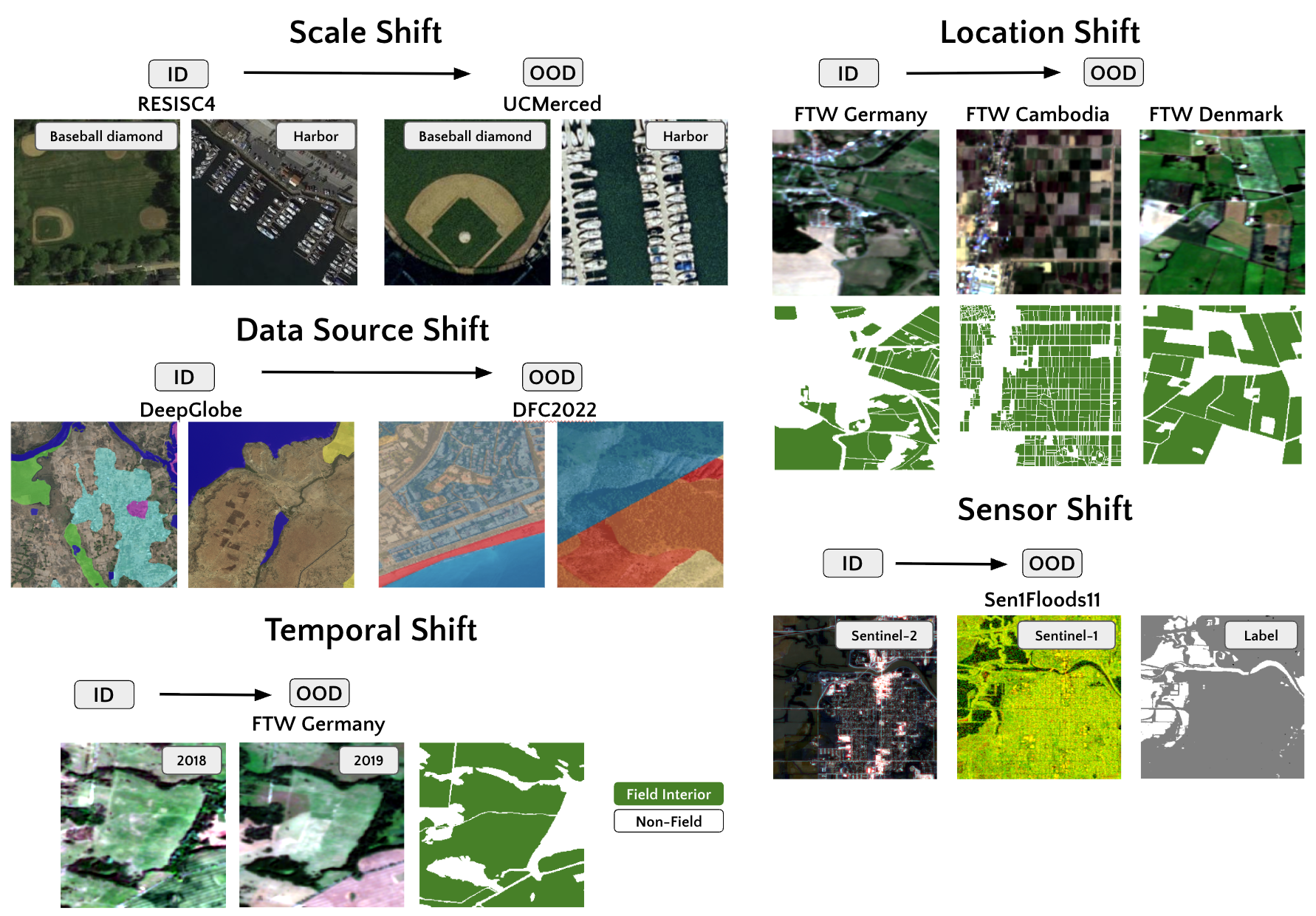}
    \caption{Representative samples of EarthShift distribution shifts. ID represents in-distribution data, OOD represents out-of-distribution data that exhibit a "shift" from the ID data.}
    \label{fig:shifts}
\end{figure*}

 Previous work has shown machine learning (ML) models for Earth Observation (EO) applications to be extremely brittle to out-of-distribution contexts \cite{tuia2016domain}. Despite this, most benchmarks evaluating ML  models for satellite data (SatML) test generalization to new samples that are closely related to the in-distribution training set, and do not account for the unique distribution shifts common to remote sensing data. When SatML models are deployed in real applications, differences in data distributions between data seen during training and inference time are common. In a changing climate with unpredictable shifts in ecosystems and the built environment, distributionally-robust models are vital for monitoring our changing planet \cite{rolf2024mission}. 

To date, there has been no comprehensive study to quantify the distributional robustness of SatML models for natural distribution shifts characteristic of real-world deployment scenarios, largely due to limited benchmark datasets representing such shifts. Although existing benchmarks show models progressing on state-of-the-art (SOTA) performance \cite{simumba2026geobench2performancecapabilityrethinking}, distributional robustness tells a different story of model applicability in real-world contexts, where sufficient progress has not been made. 

We present a new model benchmarking testbed, EarthShift, that enables the formal characterization and evaluation of model robustness to specific types of natural dataset shifts. We focus on five categories of distribution shifts common to remote sensing: shifts in spatial resolution, time, location, sensors, and sources (Figure \ref{fig:shifts}). EarthShift includes 11 evaluation tasks spanning these 5 shift types. We evaluated geospatial foundation models (GFMs), generic vision foundation models and fully-supervised models across two task types (classification and semantic segmentation) and two transfer protocols (full fine-tuning and frozen fine-tuning, i.e., fine-tuning with a frozen pre-trained backbone and simple task-specific head). Our benchmark results from 10,000 experiments showed several clear trends about the distributional robustness of models tested on EarthShift: 
\begin{itemize}
    \item All models are robust to temporal shifts but not geographic, scale, sensor, and source shifts.
    \item GFMs have similar (or sometimes worse) robustness compared to vision foundation models (e.g., CLIP or ImageNet-pretrained models), despite domain-specific pretraining.
    \item Neither GFMs or VFMs outperform traditional supervised models (randomly initialized ResNets or ViTs) in terms of distributional robustness.
    \item Model robustness does not improve, and sometimes worsens, when encoders are fine-tuned.
\end{itemize}

EarthShift provides a flexible framework to evaluate distributional robustness gaps and will encourage the development of models that generalize to diverse out-of-distribution conditions, ultimately improving GFM deployability in real-world applications.

\section{Related Work}
Most current research studying model robustness to distribution shifts is limited to synthetic shifts that transform the sample covariates of a source dataset using, for example, image corruptions \cite{hendrycks2019benchmarking, geirhos2018generalisation}, adversarial examples \cite{madry2017towards} or other augmentations \cite{geirhos2018imagenet, huang2017arbitrary}. Extensive literature exists on improving robustness through domain adaptation \cite{ganin2016domain, sun2016deep} and invariance learning \cite{arjovsky2019invariant}, however, these methods assume access to target domain data, which may not be available in EO deployment scenarios. Test-time adaptation \cite{wang2021tent} offers promise by adapting during deployment without labeled target data, though its effectiveness for EO distribution shifts remains unexplored. 

While there has been some effort to create realistic benchmarks representing natural distribution shifts for computer vision such as WILDS \cite{koh2021wilds}, ImageNetV2 \cite{shankar2020evaluating}, or ObjectNet \cite{barbu2019objectnet}, there is still a lack of out-of-distribution (OOD) benchmark datasets that capture natural distribution shifts that emulate those likely to be encountered in deployed SatML settings \cite{sachdeva2024distribution}. Datasets capturing natural distribution shifts are particularly important because robustness to synthetic dataset shifts is not representative of robustness to natural distribution shifts \cite{taori2020measuring}, and out of distribution scenarios are constantly encountered in SatML contexts. For example, \cite{benson2020assessing} showed that the accuracy of models trained to detect post-disaster building damage in remote sensing images dropped by up to 30\% when evaluated on a new disaster site. Similarly, agriculture field delineation models fail in new countries with different farming and climate patterns than in training data \cite{kerner2023multi}. 

Existing literature has not explored a holistic view of multiple distribution shift settings for geospatial foundation models. Previous work has assessed one type of distribution shift at a time, usually only evaluating a few models. For example, \cite{tamazyan2025geocrossbench} investigated robustness to sensor shift in remote sensing data and showed that RS-specific foundation models still have a lot of room for improvement to significantly outperform general-purpose models on cross-satellite generalization. \cite{sachdeva2024distribution} evaluated robustness of foundation models to geographic distribution shifts for tree detection in India. They found that some foundation models (SAM and DINO) were slightly more robust, but “not enough to justify their increased scale and cost.” \cite{benson2020assessing} investigated robustness in building damage detection, showing that supervised vision models generalized poorly to new locations and types of natural disasters, even when domain adaptation techniques were applied.

\section{EarthShift benchmark}
\subsection{Measuring distributional robustness}
The goal of EarthShift is not to evaluate ID or OOD performance in isolation but to measure robustness of models across distribution shifts (ID to OOD datasets). We measure distributional robustness using Taori et al.~\cite{taori2020measuring}'s definition of effective robustness, which is distinct from standard performance metrics used in classification and semantic segmentation.
We evaluate models on two test sets: the "in distribution" (ID) test set, which consists of samples from the same distribution as the fine-tune set, and the "out of distribution" (OOD) test set, which consists of samples that are distributionally shifted from the ID data in some way (temporal, geographic, sensor/modality, scale, and source shift). 

To compare distributional robustness between two models, one approach is to rank the models by their performance on the OOD test set. However, this method does not disentangle a model's robustness from its ID performance, and conflates robustness with overall capability. Consider three models and their scores: (a) ID: 90\%, OOD: 70\% (-20\% change), (b) ID: 15\%, OOD: 10\% (-5\% change), and (c) ID: 60\%, OOD: 60\% (0\% change). Ranking by absolute difference in OOD-ID would suggest model (b) is more robust than model (a), when in reality model (b) performs poorly everywhere. 

The effective robustness framework from \cite{taori2020measuring} addresses this by measuring performance relative to a baseline that accounts for ID capability. Effective robustness is defined as: 
\begin{equation}
    \rho(f) = score_{OOD}(f) - \beta(score_{ID}(f))
\end{equation}

Where $\beta$ is a function that estimates the baseline score (e.g., accuracy, mIoU, F1) on the OOD test set based on model $f$'s score on the ID test set. We compute $\beta$ by fitting a linear regression model to the OOD vs. ID scores  from all models evaluated on all datasets within one shift type (see the dashed black lines in Figure \ref{fig:dist-shift}). For reference, we provide the absolute ID and OOD scores of all models on all tasks, as well as their absolute difference, in Supplementary Materials. 

\subsection{Distribution Shifts}

\begin{table*}[t]
\centering
\caption{Dataset pairs grouped by distribution shift type in EarthShift (11 test cases).}
\label{tab:data}
\footnotesize 
\setlength{\tabcolsep}{3pt} 
\begin{tblr}{
  width = \textwidth,
  colspec = {
    l % Shift
    l % Task
    X[1.2,l,m] % ID Dataset 
    X[1.2,l,m] % OOD Dataset 
    c % Train
    c % Val
    c % ID Test
    c % OOD Test
  },
  row{1} = {font=\bfseries, c},
  % Row spans (r=X) must match the actual number of rows per category
  cell{2}{1} = {r=1}{l}, % Data Source (1 row)
  cell{3}{1} = {r=2}{l}, % Scale (2 rows)
  cell{5}{1} = {r=3}{l}, % Temporal (3 rows - updated from 2)
  cell{8}{1} = {r=2}{l}, % Geographic (2 rows)
  cell{10}{1} = {r=3}{l}, % Sensor (3 rows)
  hline{1,2,13} = {0.08em}, % Top, Header, and Bottom (updated from 12)
  hline{3,5,8,10} = {0.03em}, % Separators between shift types (updated from 7,9)
}
Shift       & Task        & ID Dataset                & OOD Dataset                & Train & Val  & ID Test & OOD Test \\
Data Source & SemSeg      & DeepGlobe                 & DFC2022                    & 539   & 75   & 175     & 175      \\
Scale       &  Class & RESISC45                  & UCMerced                   & 7915  & 2726 & 2659    & 1260     \\
            &  Class & UCMerced                  & RESISC45                   & 420    & 420   & 1260      & 2659       \\
Temporal    & SemSeg      & FTW SA (A) & FTW SA (B) & 590   & 72   & 85      & 85       \\
            & SemSeg      & FTW Germany (A)   & FTW Germany (B)          & 413   & 60   & 200     & 700      \\
            & SemSeg      & FTW All (A) & FTW All (B) & 49,333  & 7041   & 14,110   & 14,110       \\
Geographic  & SemSeg      & FTW Germany               & FTW Cambodia              & 612   & 60   & 700     & 50       \\
            & SemSeg      & FTW Germany               & FTW Denmark               & 612   & 60   & 700     & 663      \\
Sensor      &  Class & m-EuroSat RGB & m-EuroSAT RGE1 & 3240  & 996  & 996     & 996      \\
            & Mutli-class & BigEarthNetv2 S2  & BigEarthNetv2 S1  & 20,000    & 4000   & 4000     & 4000      \\
            & SemSeg      & Sen1Floods11 S2   & Sen1Floods11 S1   & 252    & 89   & 90      & 90      
\end{tblr}
\end{table*}

EarthShift measures distribution shifts in five categories: spatial scale shifts, temporal shifts, geographic shifts, sensor shifts and data source shifts illustrated in Figure \ref{fig:shifts} with dataset pairs summarized in Table \ref{tab:data}. Models are fine-tuned on the defined ID data and evaluated on both a held-out test set from ID data and a held-out test set from the defined OOD dataset. We leveraged datasets from the EarthNets database \cite{xiong2022earthnets}, GeoBench \cite{lacoste2023geobenchfoundationmodelsearth}, GeoBenchv2 \cite{simumba2026geobench2performancecapabilityrethinking} and Zenodo to identify groups of datasets appropriate for the same task type and overlapping classes.

\subsection{Scale shift}
The scale shift scenario assesses a model's robustness to changes in image resolution between datasets sharing the same classes and geographic coverage. In practice, this shift frequently occurs when models trained on freely available satellite imagery (typically coarser resolution due to coverage-resolution tradeoffs) are applied to  proprietary high-resolution data.

%\noindent
\paragraph{RESISC45-UCMerced scene classification.} We perform two versions of the scale shift scenario. First, we selected the RESISC45 (REmote Sensing Image Scene Classification) dataset \cite{7891544} as our ID data source and the UCMerced \cite{yang2010bag} land cover dataset as our OOD data source. Second, we reverse the order, using UCMerced as ID and RESISC45 as OOD. RESISC45 contains 31500 RGB images, covering 45 scene classes with 700 images in each class at a resolution between 0.2-30m. The UCMerced Land Use dataset is an RGB dataset of 0.3m resolution covering 21 classes of land use in 2202 images. We retained only those samples that contained one of the 19 classes shared between the two datasets: airplane, baseball diamond, beach, chaparral, dense residential, forest, highway, golf course, harbor, intersection, medium residential, mobile home park, overpass, parking lot, river, runway, sparse residential, storage tank, and tennis court.

\subsection{Temporal Shift}
The temporal shift scenario assesses a model’s robustness to changes in time between images from the same location. Applying models to future years or different seasons not covered in the training set is a ubiquitous challenge in SatML. This setting is particularly important when considering GFMs in agricultural contexts, where crops vary in their spectral signatures throughout the growing season, and visual field characteristics change. We used the Fields of The World \cite{kerner2024fieldsworldmachinelearning} (FTW) dataset to test model robustness between periods of the agriculture season in the same locations. FTW is a benchmark dataset for agricultural field segmentation, containing instance and semantic segmentation masks paired with multi-date, multi-spectral (RGB-NIR) Sentinel-2 satellite images. The FTW dataset provides natural groupings for evaluating shifts in geography and time. The complete dataset includes 24 countries, which are organized into country-level training, validation, and test subsets. FTW provides images from two dates for each sample, referred to as Window A and Window B, which correspond roughly to the planting and harvesting seasons for each location. We evaluate season-to-season performance shift using the global dataset (all 24 countries) as well as a region-specific subset (FTW-South Africa). We also evaluated a year-to-year shift using the FTW-Germany subset, which has training data from 2018 and test data from 2019. 

\paragraph{FTW Season-to-Season}
Samples from Window A per country are used in the ID dataset, samples from Window B per country are used in the OOD dataset. Full details of corresponding time period for each window and country are available in \cite{kerner2024fieldsworldmachinelearning}.

\paragraph{FTW-South Africa Season-to-Season}
In the season-to-season shift for South Africa, the images in windows A and B are drawn from non-overlapping time periods: (A) 2018/07/01 - 2018/09/30, (B) 2018/04/01 - 2018/06/30. We selected window A as our ID set and window B as our OOD set.

\paragraph{FTW-Germany Year-to-Year}
In the year-to-year shift for Germany, the samples from windows A and B are drawn from the following periods in the training (2018) and test (2019) sets respectively: (A) 2019/07/01 - 2019/09/30, 2018/07/01 - 2018/09/30, (B) 2019/04/01 - 2019/06/01, 2018/04/01 - 2018/06/01. We used the 2018 data as our ID set and the test period of 2019 as our OOD set.

\subsection{Geographic Shift}
The geographic shift setting assesses a model’s ability to generalize between locations. We used the country subsets in FTW to create two geographic shift scenarios. We chose these pairings because of their differences in geographic location, field boundary size, and shape across regions.

\paragraph{FTW Germany-Cambodia}
In this task, we used Germany as the ID set and Cambodia as the OOD set. Germany has much larger agricultural plots on average relative to Cambodia, and both countries are on different continents with very different agricultural seasons.

\paragraph{FTW Germany-Denmark}
In this task, we again used Germany as the ID set but chose a different country, Denmark, as the OOD set. Germany and Denmark share the same continent but are relatively far apart with different seasonal and agroecological characteristics. 

\subsection{Sensor Shift} 
The sensor shift scenario tests a model's robustness to radiometry changes between image inputs. Foundation models have shown performance degradation when processing imagery with different spectral characteristics which is detrimental particularly in geospatial contexts where real-world applications can require models to summarize data from various sensors \cite{tamazyan2025geocrossbench}. We explore distribution shift between sensor types from Sentinel-2 to Sentinel-1 (S2-S1) and a partial band overlapping scenario whereby OOD data contains a subset of the channel radiometry of the ID data.

\paragraph{BigEarthNetv2 S2-S1 multi-label classification}
BigEarthNet V2 \cite{clasen2024reben} is a multi-label classification dataset containing 19 classes of Sentinel 1 and 2 imagery over Europe. We use the Sentinel-2 channels as the ID dataset and the Sentinel-1 channels as the OOD dataset. The labels are derived from the CORINE Land Cover labels.

\paragraph{Sen1Floods11 S2-S1 segmentation} Sen1Floods11 \cite{9150760} is a surface water dataset of labelled permanent water and flood water. The dataset consists of 4,831 512x512 chips covering 120,406 km2, spanning 6 continents of the world across 11 flood events. We utilize the hand-labelled flood event samples, with Sentinel-2 data used as the ID samples and Sentinel-1 data as the OOD samples. This scenario is included to investigate robustness in the disaster relief domain of remote sensing applications.

\paragraph{m-EuroSat RGB-RGE1}
We drew inspiration from GeoCrossBench, and tested sensor shift robustness in a partial band overlap setting. In this scenario, samples in our in-distribution dataset contain RGB (B02, B03, B04) channels from Sentinel-2. Out-of-distribution test samples contain R, G, and Vegetation Red Edge channels (B02, B03, B05 or ``RGE1''). We used the 0.2x training partition of the m-EuroSat dataset from Geo-Bench \cite{lacoste2023geobenchfoundationmodelsearth}. 

\subsection{Data source shift}
The data source shift setting challenges the assumption that models should perform similarly on datasets that are visually similar. The setting is inspired by the generalization gap shown between ImageNet and ImageNet-v2 \cite{recht2019imagenetclassifiersgeneralizeimagenet}---these datasets were created using similar strategies at two different times and were indistinguishable to humans, yet ImageNet-trained models suffered significant performance drops when tested on ImageNet-v2. The data source shift scenario tests a model's robustness to shifts between dataset sources that share the same task, sensor channels, classes, spatial resolution and geographic coverage, but are generated using different methodologies. 

%\noindent
\paragraph{DeepGlobe-DFC2022 semantic segmentation.} We selected the DeepGlobe Land Cover Classification Challenge \cite{DeepGlobe18} dataset as our ID data source and the DFC2022 2022 IEEE GRSS Data Fusion Contest dataset \cite{dfc2022_dataset} as our OOD data source. DeepGlobe contains 803 satellite RGB images with size 2448x2448 from the DigitalGlobe satellite at 0.5 m/pixel resolution and 7 segmentation classes. The DFC2022 dataset consists of RGB aerial images at 0.5 m/pixel spatial resolution with size 2000x2000 and 15 segmentation classes. We mapped the following classes in common between the DeepGlobe and DFC2022, respectively: urban land: urban fabric, forest land: forests, water: water, barren land: open spaces with little to no vegetation. We ensured all samples used in training, validation and testing contained at least one of these four overlapping classes in the label mask to adequately calculate mIoU as our performance metric. %Both the DFC2022 and DeepGlobe datasets provide labels only for their author-defined train splits. Therefore, we split the data as follows: we chose 550 samples for the train split, 75 samples for validation, and 175 samples for the in-distribution test set (DeepGlobe). For the OOD test set, we selected 175 samples from DFC2022 to maintain the same size between the two test sets. 

\begin{figure*}[htbp]
    \centering
    \includegraphics[width=\textwidth]{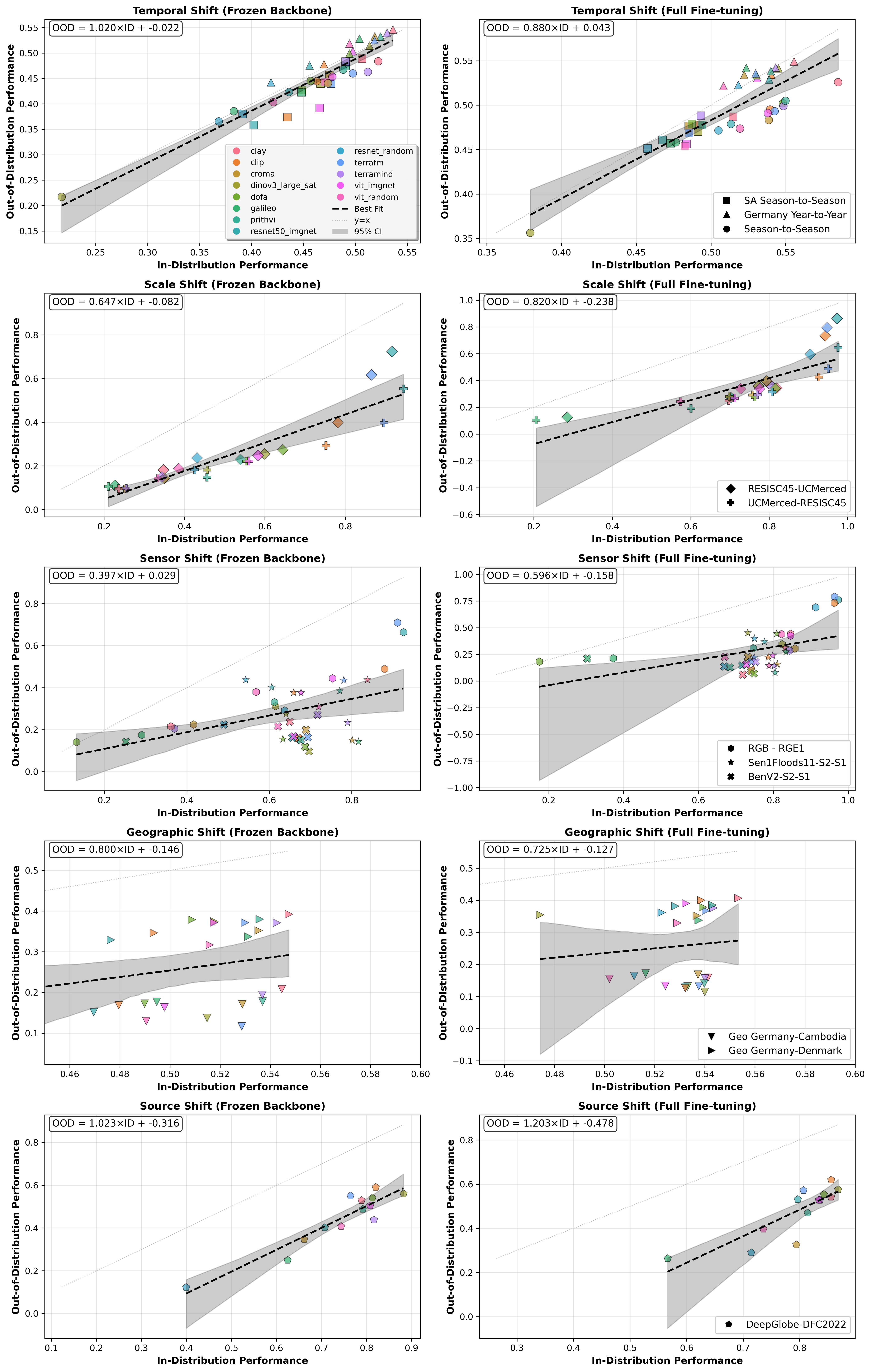}
    \caption{Effective distribution shift analysis for (left) frozen backbone and (right) full fine-tuning. Each point represents a model-task combination, with colours denoting model types. The dashed y=x line represents perfect robustness (no performance drop). The black dashed line represents the line of best fit calculated as OOD=m*ID+b. Points indicate mean over 5 random seeds.}
    \label{fig:dist-shift}
\end{figure*}

\section{Methods}
\subsection{Model architectures}
%Table \ref{tab:models} summarizes the characteristics of the models currently supported in EarthShift. 

\paragraph{Geospatial Foundation Models (GFMs)} We selected 8 open-source models pre-trained on global archives of remote sensing data: DOFA \cite{xiong2024neural}, CROMA \cite{fuller2023CROMAremotesensingrepresentations}, Clay \cite{clay_foundation_model_2024}, Prithvi V2.0 \cite{szwarcman2025prithvieo20versatilemultitemporalfoundation}, TerraFM \cite{danish2025TerraFM}, DINOv3 \cite{simeoni2025DINOv3} (pre-trained on the SAT-493M dataset), Galileo \cite{tseng2025galileolearningglobal}, and TerraMind \cite{jakubik2025terramindlargescalegenerativemultimodality}.

%\noindent
\paragraph{Generic Vision Foundation Models (VFMs)} We included multiple vision foundation models to compare distributional robustness to our selected GFMs. We included the ResNet50 \cite{he2015deepresiduallearningimage} and Vision Transformer ViT-b-16 \cite{dosovitskiy2021imageworth16x16words} architectures pre-trained on ImageNet. We also included CLIP \cite{Radford2021LearningTV}. 

\paragraph{Fully-Supervised Models}
To evaluate the benefit of transfer learning from ImageNet or remote sensing pretrained models, we also included randomly-initialized ResNet50 and ViT-b-16 models.

\subsection{Training setup}
We trained models under two settings: (i) full fine-tuning, where all parameters of the pre-trained model and the task-specific head are updated; and (ii) fine-tuning with frozen backbone, where only the parameters of a newly added task-specific head are trained.

\paragraph{Input channel adaptation}
The expected input size and channel number for each model varies. To adapt models to various input channel sizes, we implemented a learnable 1x1 convolution channel adapter similar to \cite{tamazyan2025geocrossbench}. We resized all samples to match the input size that was used during pre-training for each model architecture (see Supplementary Materials for additional details).  

\paragraph{Task adaptation}
For semantic segmentation tasks, we add a simple decoder head consisting of a convolution - ReLU - convolution sequential module. For scene classification tasks we add a linear classifier as the final layer. We note that the goal of EarthShift is not to maximize each model's performance by finding the optimal task adaptation decoder, so evaluating permutations of the final task-specific classification or decoding heads is not practical or in scope for this work. Since our goal is to assess distributional robustness across tasks, we chose a simple task decoder that introduces minimal hyperparameters influencing the respective model performance. 
% \noindent
\paragraph{Learning rate} We swept learning rates similar to \cite{tamazyan2025geocrossbench, tseng2025galileolearningglobal} over the sets $\{1, 3, 6\} \times \{10-5, 10-4, 10-3\}$ for full fine-tuning and $\{1, 3, 5\} \times \{10-4, 10-3, 10-2\}$ for fine-tuning with frozen backbone. We performed learning rate sweeps for all models for at least one task within each type of distribution shift: scale shift with RESISC45-UCMerced and dataset source shift with DeepGlobe-DFC2022, sensor shift with m-Eurosat RGB-RGE1, Sen1Floods11, and BigEarthNetv2, geographic shift with FTW Germany-Cambodia, and temporal shift with FTW South Africa seasonal. We used the best learning rate from the model validation score and implemented this learning rate in any other tasks of the same shift type (i.e., the identified best learning rate from the temporal shift learning rate sweep was used across temporal shift tasks).

% \noindent
\paragraph{Training} We fine-tuned each model under two conditions for the frozen backbone and full-fine settings. Our fine-tuning with frozen backbone setup trains for 40 epochs and our full fine-tuning setup trains for 50, both with a batch size of 32, AdamW optimizer \cite{loshchilov2019decoupled} and a learning-rate scheduler featuring a 20-epoch linear warmup phase followed by a cosine decay. We selected epoch fine-tuning length based on observed convergence behaviour of the train and validation loss curves during our learning rate sweep experiments across tasks. We trained all models using cross-entropy loss for segmentation and classification tasks and binary cross entropy loss for multi-label tasks. 

\paragraph{Evaluation} We evaluated performance for classification tasks using accuracy, multi-label classification using micro-F1 score, and semantic segmentation using the mIoU. We applied channel-wise mean and standard deviation normalization, using Sentinel-2 or ImageNet metrics where appropriate for each dataset (i.e., RGB datasets used ImageNet statistics and Sentinel-2 datasets used Sentinel-2 metrics as defined by Clay \cite{clay_foundation_model_2024}).

\section{Results}

\begin{table*}[t]
\centering
\footnotesize
\setlength{\tabcolsep}{4pt}
\caption{Effective robustness score per model (rows) and shift type (columns).
Cells are colored from green (best) to red (worst); the highest score in each shift column is bolded. Models are ordered top-to-bottom by average $\rho$ across shifts. Per-model, task robustness scores in Supplementary Materials. Avg $\rho$ column is the average effective robustness (\%) across all shifts. No single model dominates: in full fine-tuning the top three models each take rank 1 in at least one shift but never in more than two. Surprisingly, in both the full fine-tuning and frozen backbone settings, the ImageNet ResNet50 (R50-imgnet) is the most robust model overall, and the randomly-initialized ResNet50 (R50-rnd) -- which uses random convolutions in the encoder -- is second in the frozen setting.}
\label{tab:model_ranks_eff}
\begin{subtable}{0.49\textwidth}
\centering
\caption{Full Fine-Tuning}
\label{tab:model_ranks_full_eff}
\begin{tabular}{l@{\hskip 4pt}ccccc@{\hskip 6pt}c}
\toprule
\textbf{Model} & \textbf{Sensor} & \textbf{Geog.} & \textbf{Temp.} & \textbf{Scale} & \textbf{Src.} & \textbf{Avg $\rho$} \\
\midrule
R50-imgnet & \rB{9.5}  & \rD{-0.1}  & \rD{0.1}  & \rA{19.4}  & \rC{5.2}  & \rA{6.6}  \\
Galileo    & \rB{9.2}  & \rD{0.0}  & \rA{1.1}  & \rB{15.3}  & \rB{6.0}  & \rA{6.1}  \\
TerraFM    & \rC{6.9}  & \rG{-1.3} & \rD{-0.1}  & \rB{10.1}  & \rA{8.0}  & \rB{4.2}  \\
R50-rnd    & \rA{10.5}  & \rA{1.5}  & \rB{0.5}  & \rD{-0.8}  & \rF{-9.1} & \rB{2.3}  \\
CLIP       & \rD{1.8}  & \rC{0.2}  & \rE{-0.2}  & \rC{5.1}  & \rB{6.9}  & \rB{2.1}  \\
DOFA       & \rC{4.7}  & \rF{-0.7} & \rB{0.3}  & \rG{-10.0} & \rC{1.9}  & \rC{-0.4}  \\
DINOv3     & \rD{-0.2}  & \rF{-0.6} & \rE{-0.2} & \rD{-4.6}  & \rD{1.0}  & \rD{-1.0}  \\
ViT-rnd    & \rD{-2.3}  & \rE{-0.5} & \rF{-0.3} & \rC{-0.5}  & \rE{-1.0} & \rD{-1.0}  \\
ViT-imgnet & \rE{-3.9}  & \rB{0.6}  & \rF{-0.6} & \rE{-6.7} & \rD{0.4}  & \rE{-2.3}  \\
Clay       & \rE{1.8} & \rB{1.3}  & \rG{-0.8} & \rF{-8.6} & \rE{-0.9}  & \rE{-2.9}  \\
TerraMind  & \rG{-11.5} & \rC{0.1}  & \rC{0.2}  & \rF{-7.1} & \rD{-0.01}  & \rF{-4.4}  \\
Prithvi-2.0    & \rF{-11.2} & \rE{-0.3}  & \rD{-0.1}  & \rF{-6.4} & \rF{-3.1} & \rF{-4.6} \\
CROMA      & \rF{-8.8} & \rD{-0.2}  & \rC{0.2}  & \rD{-5.3}  & \rG{-15.1} & \rG{-4.7} \\
\bottomrule
\end{tabular}
\end{subtable}
\hfill
\begin{subtable}{0.49\textwidth}
\centering
\caption{Frozen Backbone}
\label{tab:model_ranks_frozen_eff}
\begin{tabular}{l@{\hskip 4pt}ccccc@{\hskip 6pt}c}
\toprule
\textbf{Model} & \textbf{Sensor} & \textbf{Geog.} & \textbf{Temp.} & \textbf{Scale} & \textbf{Src.} & \textbf{Avg $\rho$} \\
\midrule
R50-imgnet & \rB{9.1}  & \rC{0.8}  & \rD{0.1}  & \rA{11.9}  & \rD{-0.5}  & \rA{4.3}  \\
R50-rnd    & \rB{6.8}  & \rA{4.7}  & \rA{1.7}  & \rD{1.5}  & \rC{3.0}  & \rA{3.5}  \\
TerraFM    & \rA{9.2}  & \rF{-3.3} & \rF{-1.1} & \rC{2.0}  & \rA{8.5}  & \rB{3.1}  \\
Clay       & \rC{2.3}  & \rC{0.9}  & \rE{-0.4}  & \rB{3.2}  & \rB{3.8}  & \rB{2.0}  \\
CLIP       & \rD{2.1}  & \rB{1.4}  & \rF{-1.1} & \rG{-6.8} & \rB{6.7}  & \rC{0.5}  \\
ViT-rnd    & \rD{2.0}  & \rG{-3.3} & \rB{0.6}  & \rC{1.5}  & \rF{-3.6} & \rC{-0.6}  \\
ViT-imgnet & \rC{2.4}  & \rD{0.8}  & \rG{-1.8} & \rF{-5.3} & \rD{-0.5}  & \rD{-0.9}  \\
Galileo    & \rE{-5.5} & \rE{-0.7}  & \rB{1.4}  & \rB{4.9}  & \rF{-7.3} & \rE{-1.4}  \\
Prithvi-2.0 & \rD{-1.3}  & \rD{-0.4} & \rD{0.2}  & \rE{-5.1} & \rD{-0.6}  & \rE{-1.4}  \\
DOFA       & \rF{-8.3} & \rB{2.3}  & \rC{0.3}  & \rF{-5.8} & \rC{2.5}  & \rE{-1.8}  \\
CROMA      & \rG{-9.0} & \rF{-1.8} & \rD{0.03}  & \rD{0.7}  & \rE{-1.4} & \rF{-2.3}  \\
TerraMind  & \rE{-4.0}  & \rD{-0.3}  & \rE{-0.4} & \rD{1.2}  & \rG{-8.1} & \rF{-2.3}  \\
DINOv3     & \rF{-5.9} & \rE{-1.1} & \rC{0.5}  & \rE{-4.1}  & \rE{-2.6}  & \rG{-2.6}  \\

\bottomrule
\end{tabular}
\end{subtable}
\label{tab:rank}
\end{table*}

\paragraph{GFMs show no robustness advantage over generic vision models or random baselines.}  The average effective robustness across all tasks is $\sim0\%$. In all shifts (except geographic), models tend to cluster at the baseline, indicating no genuine robustness improvements beyond what their ID performance predicts. Baseline slopes varied by shift type: temporal  (m=1.02, 0.88 for frozen, full, showing near-perfect robustness), geographic (m=0.8, 0.72 for frozen, full), scale (m=0.65, 0.82 for frozen, full), sensor (m=0.39, 0.59 for frozen, full, showing severe degradation), and source (m=1.0, 1.2 for frozen, full).

\paragraph{All models show strong robustness to temporal shifts.}
We find little to no performance drop across models with a baseline slope $\sim1$ showing that models are robust. However, this reflects shift difficulty; all models show similar robustness and there is no advantage of GFMs to VFMs or fully-supervised models. We therefore interpret this as both an example of robust generalization in models as well as a flag for including additional temporal shift samples for future work. 

\paragraph{Models are least robust to sensor shifts}
All models suffer from severe degradation under sensor shift distribution and high heterogeneity (Figure \ref{fig:dist-shift}). GFMs perform the worst on average.

\paragraph{Bigger geographic shifts result in worse robustness}
When trained on FTW-Germany, all models show significantly worse generalization to Cambodia compared to Denmark. There are clear spatial differences between cropping patterns in Germany and Cambodia (countries on different continents with different dominant crops), whereas there are more similarities (same continent, similar crop types) between Germany and Denmark (Figure \ref{fig:shifts}).

\paragraph{Models with strong ID performance lose substantial robustness under scale shift} In full fine-tuning, $\beta$ has a higher slope (0.82) but a much more negative intercept (−0.24) than the frozen backbone setting. We see that even models achieving strong ID performance (0.95) experience significant OOD degradation, with the average model losing ~43\% of its ID performance, and the full range of models (confidence band) showing OOD losses ranging from $\sim$32-63\%.

\section{Discussion}
The EarthShift benchmark evaluates 13 foundation model architectures—including 8 state-of-the-art Geospatial Foundation Models (GFMs)—to characterize their distributional robustness across diverse GeoAI contexts. While we have not included an exhaustive survey of every available model, our selection captures the SOTA architectural and pre-training paradigms currently dominant in the field. Consequently, we expect the observed trends to serve as a reliable proxy for the broader landscape of GFMs and discuss here key observations.

\subsection{Evaluated models lack distributional robustness for Earth observation}
We find models lose 20-25\% of their performance when evaluated out-of-distribution on average Figure~\ref{fig:dist-shift}), with substantial variation by shift type 
(temporal: $\sim$2\% loss, sensor: up to $\sim$60\% loss) is comparable to the 14-30\% gaps reported for ImageNet to ImageNet-V2 \cite{recht2019imagenetclassifiersgeneralizeimagenet, taori2020measuring}. However, our results reveal that pre-training on Earth Observation data does not confer robustness advantages over generic vision models pre-trained on ImageNet, or full supervised models with no pre-training (Figure \ref{fig:dist-shift}). This behaviour suggests that current GFM pre-training strategies may not adequately address distributional robustness and could be overfitting to patterns, such as simplicity biases, in the training data.

\subsection{Pre-training on EO data does not help robustness}
The finding that GFMs show no robustness advantage over ImageNet models (Table \ref{tab:rank}) challenges assumptions about domain-specific pre-training. We speculate several reasons why this performance gap persists. EO pre-training datasets may contain systematic biases that ImageNet avoids through photographic diversity. For example, if most pre-training data comes from Sentinel over developed regions, models learn biases specific to these sensors and geographies that hurt generalization. ImageNet contains 1.2M images across 1000 diverse categories from varied contexts \cite{deng2009imagenet}. This diversity may result in superior robustness despite being out-of-domain to remote sensing applications. 

\subsection{Limitations and Future Work}
EarthShift is the first testbed and collection of datasets to enable systematic benchmarking of distributional robustness. We note that EarthShift does not include an exhaustive analysis of the distribution shifts across the task scenarios, limiting our ability to make broad claims about robustness to these types of shifts in general without a more extensive experiment suite. Additional datasets would help strengthen our claims, but are unlikely to change our findings significantly, since we see consistent trends across many models and tasks. The design of EarthShift enables the flexibility to add tasks and models into the testbed for future work, and we hope this testbed will become a living resource for testing distributional robustness as new datasets and models are added in the future. 

\section{Conclusion}
EarthShift is the first comprehensive benchmark for measuring distributional robustness across realistic distribution shifts encountered in SatML applications. Through systematic evaluation of 13 models (including 8 GFMs) across 5 shift types and 11 tasks, we demonstrated that current models exhibit 20\% performance degradation on average (Figure \ref{fig:dist-shift}) to out-of-distribution. Our benchmark results reveal several key insights; most surprisingly, we found that GFMs show no robustness advantage over generic vision models or even fully-supervised baselines (for the 13 models tested), challenging assumptions about the benefits of domain-specific pretraining. We release our code, datasets and onboarding instructions for contributing to Earthshift to enable reproducible robustness evaluation and accelerate progress toward robust geospatial AI. 

%Our results highlight the urgent need for the field to prioritize model robustness alongside performance. As Earth observation models are increasingly deployed for climate monitoring, agricultural management, and disaster response \cite{rolf2024mission}, their brittleness under distribution shift poses risks for real-world applications. EarthShift provides the evaluation framework to guide future research toward models that generalize reliably across the temporal, geographic, and sensor variations inherent to Earth observation. 

\bibliographystyle{unsrt}
\bibliography{main}

%%%%%%%%%%%%%%%%%%%%%%%%%%%%%%%%%%%%%%%%%%%%%%%%%%%%%%%%%%%%
\newpage
\appendix

\section{Technical appendices and supplementary material}
\subsection{Code and Data}
Code and data for EarthShift are available: \href{URL}{https://anonymous.4open.science/r/earthshift-8736}.

\subsection{Dataset License Details}
\begin{table}[htbp]
\centering
\caption{Dataset Licenses used in EarthShift}
\begin{tabular}{l|l}
Dataset        & License                                                      \\ 
\hline
RESISC45       & CC BY-NC-SA 4.0                                              \\
UCMerced       & Public Domain                                                \\
DeepGlobe      & Non-commercial research and educational use only             \\
DFC2022        & IGN’s “licence ouverte"                                      \\
FTW            & CC BY-SA 4.0                                                 \\
Sen1Floods11   & Open Access                                                  \\
BigEarthNet v2 & Community Data License Agreement – Permissive – Version 1.0  \\
m-EuroSat      & MIT License                                                 
\end{tabular}
\end{table}

\subsection{Geospatial Foundation Model Details}
\begin{table}[htbp]
\centering
\caption{Characteristics of models compared: model name, number of backbone parameters, learning technique, and pre-training dataset.}
\small
\label{tab:model_ranks}
\begin{tblr}{
  hline{2} = {-}{},
}
Model      & Num Parameters & Learning Technique &  Pre-training Dataset \\
Galileo    &     86.5M       &    Self-supervised     &    Sentinel-1, Sentinel-2         \\
DOFA       &     111.3M      &    Self-Supervised (MAE)                   &    Sentinel-1, Sentinel-2, NAIP, Hyperspectral         \\
CROMA      &     90.3M       &   Self-Supervised (Contrastive + MAE)      &     Sentinel-1, Sentinel-2       \\
Clay       &     92.1M       &     Self-supervised (MAE)    &   Sentinel-2, Landsat, DEM        \\
Prithvi-2.0 &     303.9M       &     Self-Supervised (MAE)                  &     Landsat 8/9, Sentinel-2        \\
TerraFM    &     113M       &     Self-Supervised (MAE)                 &   SSL4EO-L, GeoMeld S1, S2, DEM, LULC         \\
DINOv3     &     303.1M       &    Self-Supervised                 &    LVD-142M + Aerial         \\
TerraMind  &     86.9M       &    Self-Supervised (Generative)                   &   Sentinel-1, Sentinel-2, DEM, LULC         \\
ResNet50   &     25.6M       &      Supervised     &  ImageNet             \\
ViT-b-16   &     86.6M       &     Supervised      &   ImageNet           \\
CLIP       &     86.2M       &    Contrastive   &     WIT / WebImageText         \\
           &            &                    &              
\end{tblr}
\label{tab:models}
\end{table}

\newpage

\subsection{Experimental Details}
We utilized the PSC Bridges-2 Supercomputer leveraging 4.7K GPU hours, fine-tuning models on the 24 2-socket Intel Xeon “Cascade Lake” v100-32 nodes with eight V100 GPUs with NVLink, each with 32GB of GPU memory. These nodes have 512GB RAM per node. This work used Bridges-2 at Pittsburgh Supercomputing Center through allocation CIS260071 from the Advanced Cyberinfrastructure Coordination Ecosystem: Services \& Support (ACCESS) program, which is supported by National Science Foundation grants 2138259, 2138286, 2138307, 2137603, and 2138296 \cite{Brown2021}. The authors acknowledge Research Computing at Arizona State University for providing HPC and storage resources that have contributed to the research results reported within this paper \cite{HPC:ASU23}. 

We summarize here the scale of experiments conducted. We began with learning rate sweeps per model, per task type for the two fine-tuning settings (frozen backbone and full fine-tuning). This includes 13 models, two fine-tuning settings across 7 shifts and 18 learning rates (9 per fine-tuning strategy), resulting in a total of \textbf{8190} runs. Each learning rate sweep was ran with 5 different seeds and the best learning rate was selected after calculating the average validation score across the seeds. The models were then retrained per task using 5 different random seeds to ensure robust and stable performance reporting. We include the best learning rate per model, fine-tuning setting in the tables below. 

\paragraph{Scene Classification} We evaluate 13 models in two fine-tuning settings across 4 classification tasks and 5 random seeds, totaling \textbf{520} classification task runs.

\paragraph{Segmentation}  We evaluate 13 models in two fine-tuning settings across 7 segmentation tasks and 5 random seeds, totaling \textbf{910} segmentation runs.

\subsection{Extended Results - Effective Robustness}
In this section, we present the extended results for each shift experiment for frozen and full fine-tuning, per model. 

% TEMPORAL SHIFT TABLE
\begin{table}[htbp]
\small
\centering
\caption{\textbf{Temporal Shift} Effective Robustness.}
\label{tab:temporal_shift_rho}
\begin{tblr}{
  colsep=2.5pt,
  cell{2}{2} = {c=13}{c},
  cell{6}{1} = {font=\bfseries},
  cell{7}{2} = {c=13}{c},
  cell{11}{1} = {font=\bfseries},
  vline{2} = {3-6,8-11}{},
  hline{1-3,7-8,12} = {-}{},
}
                     & \rotatebox{90}{Clay} & \rotatebox{90}{CLIP} & \rotatebox{90}{CROMA} & \rotatebox{90}{DINOv3} & \rotatebox{90}{DOFA} & \rotatebox{90}{Galileo} & \rotatebox{90}{Prithvi-2.0} & \rotatebox{90}{R50-imgnet} & \rotatebox{90}{R50-rnd} & \rotatebox{90}{TerraFM} & \rotatebox{90}{TerraMind} & \rotatebox{90}{ViT-imgnet} & \rotatebox{90}{ViT-rnd} \\
                     & Full Fine-tuning (OOD = 0.880×ID +4.3) &        &        &        &        &        &        &         &        &         &           &            &         \\
Season-to-Season     & -3.2                                   & -2.3 & -3.4 & -2.0 & -2.3 & -0.4 & -2.2 & -1.6  & -1.6 & -2.7  & -2.6    & -2.5     & -2.6  \\
Germany Year-to-Year & +1.7                                   & +1.6 & +3.2 & +1.9 & +2.4 & +3.8 & +2.0 & +1.2  & +2.4 & +2.6  & +2.1    & +2.0     & +3.1  \\
SA Season-to-Season  & -0.9                                   & +0.1 & +0.7 & -0.5 & +0.7 & -0.2 & 0.0 & +0.6  & +0.6 & -0.2  & +1.1    & -1.2     & -1.4  \\
Average              & -0.8                                   & -0.2 & +0.2 & -0.2 & +0.3 & +1.1 & -0.1 & +0.1  & +0.5 & -0.1  & +0.2    & -0.6     & -0.3  \\
                     & Frozen Backbone (OOD = 1.020×ID -2.2)  &        &        &        &        &        &        &         &        &         &           &            &         \\
Season-to-Season     & -2.7                                   & -0.6 & -2.1 & +1.8 & +0.1 & +1.7 & -0.9 & +0.1  & +1.2 & -2.6  & -3.7    & -1.2     & -0.4  \\
Germany Year-to-Year & +2.1                                   & +2.1 & +2.6 & +1.3 & +1.7 & +3.6 & +1.9 & +3.3  & +3.7 & +1.8  & +2.0    & +1.8     & +3.7  \\
SA Season-to-Season  & -0.6                                   & -4.8 & -0.6 & -1.4 & -0.9 & -1.3 & -0.4 & -3.0  & +0.2 & -2.5  & +0.4    & -6.1     & -1.5  \\
Average              & -0.4                                   & -1.1 & 0.0 & +0.5 & +0.3 & +1.4 & +0.2 & +0.1  & +1.7 & -1.1  & -0.4    & -1.8     & +0.6  
\end{tblr}
\end{table}

% SENSOR SHIFT TABLE
\begin{table}[htbp]
\small
\centering
\caption{\textbf{Sensor Shift} Effective Robustness.}
\label{tab:sensor_shift_rho}
\begin{tblr}{
  colsep=2.5pt,
  cell{2}{2} = {c=13}{c},
  cell{6}{1} = {font=\bfseries},
  cell{7}{2} = {c=13}{c},
  cell{11}{1} = {font=\bfseries},
  vline{2} = {3-6,8-11}{},
  hline{1-3,7-8,12} = {-}{},
}
                     & \rotatebox{90}{Clay} & \rotatebox{90}{CLIP} & \rotatebox{90}{CROMA} & \rotatebox{90}{DINOv3} & \rotatebox{90}{DOFA} & \rotatebox{90}{Galileo} & \rotatebox{90}{Prithvi-2.0} & \rotatebox{90}{R50-imgnet} & \rotatebox{90}{R50-rnd} & \rotatebox{90}{TerraFM} & \rotatebox{90}{TerraMind} & \rotatebox{90}{ViT-imgnet} & \rotatebox{90}{ViT-rnd} \\
                     & Full Fine-tuning (OOD = 0.596×ID -15.8) &        &        &        &        &        &        &         &         &         &           &         &         \\
RGB → RGE1           & +9.6                                    & +31.8  & -4.7   & +1.2   & +23.6  & +14.9  & +2.4   & +34.1   & +30.6   & +37.3   & -5.8      & +8.0    & +10.9   \\
BigEarthNet S2 → S1  & -20.9                                   & -17.6  & -5.2   & -19.3  & -21.6  & +19.0  & -12.0  & -12.2   & -10.4   & -10.1   & -11.0     & -11.9   & -1.4    \\
Sen1Floods11 S2 → S1 & -2.2                                    & -8.7   & -16.7  & +17.6  & +12.1  & -6.3   & -24.1  & +6.4    & +11.1   & -6.5    & -17.7     & -7.8    & -16.5   \\
Average              & -4.5                                    & +1.8   & -8.8   & -0.2   & +4.7   & +9.2   & -11.2  & +9.5    & +10.5   & +6.9    & -11.5     & -3.9    & -2.3    \\
                     & Frozen Backbone (OOD = 0.397×ID +2.9)   &        &        &        &        &        &        &         &         &         &           &         &         \\
RGB → RGE1           & +4.3                                    & +11.1  & +3.1   & +3.9   & +6.0   & +3.1   & +5.9   & +26.8   & +1.0    & +31.9   & +3.0      & +11.6   & +12.6   \\
BigEarthNet S2 → S1  & -5.0                                    & -13.4  & -10.3  & -20.9  & -18.4  & +1.5   & -14.6  & -12.7   & +0.1    & -14.0   & -4.2      & -12.3   & -6.0    \\
Sen1Floods11 S2 → S1 & +7.6                                    & +8.6   & -19.7  & -0.8   & -12.5  & -21.0  & +5.0   & +13.2   & +19.3   & +9.7    & -10.8     & +7.8    & -0.6    \\
Average              & +2.3                                    & +2.1   & -9.0   & -5.9   & -8.3   & -5.5   & -1.3   & +9.1    & +6.8    & +9.2    & -4.0      & +2.4    & +2.0    
\end{tblr}
\end{table}

% GEOGRAPHIC SHIFT TABLE
\begin{table}[htbp]
\small
\centering
\caption{\textbf{Geographic Shift} Effective Robustness. Germ = Germany, Den = Denmark, Camb = Cambodia.}
\label{tab:geographic_shift_rho}
\begin{tblr}{
  colsep=2.5pt,
  cell{2}{2} = {c=13}{c},
  cell{5}{1} = {font=\bfseries},
  cell{6}{2} = {c=13}{c},
  cell{9}{1} = {font=\bfseries},
  vline{2} = {3-5,7-9}{},
  hline{1-3,6-7,10} = {-}{},
}
                   & \rotatebox{90}{Clay} & \rotatebox{90}{CLIP} & \rotatebox{90}{CROMA} & \rotatebox{90}{DINOv3} & \rotatebox{90}{DOFA} & \rotatebox{90}{Galileo} & \rotatebox{90}{Prithvi-2.0} & \rotatebox{90}{R50-imgnet} & \rotatebox{90}{R50-rnd} & \rotatebox{90}{TerraFM} & \rotatebox{90}{TerraMind} & \rotatebox{90}{ViT-imgnet} & \rotatebox{90}{ViT-rnd} \\
                   & Full Fine-tuning (OOD = 0.725×ID -12.7) &        &        &        &        &        &        &         &         &         &           &         &         \\
Germ → Camb & -10.7                                   & -13.2  & -9.4   & -15.0  & -12.8  & -7.5   & -12.4  & -12.9   & -8.0    & -13.0   & -10.8     & -12.0   & -8.2    \\
Germ → Den  & +13.3                                   & +13.7  & +9.0   & +13.8  & +11.5  & +7.5   & +11.9  & +12.6   & +11.0   & +10.3   & +10.9     & +13.2   & +7.3    \\
Average            & +1.3                                    & +0.2   & -0.2   & -0.6   & -0.7   & 0.0   & -0.3   & -0.1    & +1.5    & -1.3    & +0.1      & +0.6    & -0.5    \\
                   & Frozen Backbone (OOD = 0.800×ID -14.6)  &        &        &        &        &        &        &         &         &         &           &         &         \\
Germ → Camb & -8.2                                    & -6.9   & -10.6  & -12.9  & -7.3   & -7.2   & -10.6  & -7.8    & +0.8    & -16.0   & -9.0      & -8.9    & -11.7   \\
Germ → Den  & +10.0                                   & +9.8   & +7.0   & +10.6  & +11.8  & +5.9   & +9.8   & +9.4    & +8.7    & +9.4    & +8.3      & +10.4   & +5.0    \\
Average            & +0.9                                    & +1.4   & -1.8   & -1.1   & +2.3   & -0.7   & -0.4   & +0.8    & +4.7    & -3.3    & -0.3      & +0.8    & -3.3    
\end{tblr}
\end{table}

% SCALE SHIFT TABLE
\begin{table}[htbp]
\small
\centering
\caption{\textbf{Scale Shift} Effective Robustness.}
\label{tab:scale_shift_rho}
\begin{tblr}{
  colsep=2.5pt,
  cell{2}{2} = {c=13}{c},
  cell{5}{1} = {font=\bfseries},
  cell{6}{2} = {c=13}{c},
  cell{9}{1} = {font=\bfseries},
  vline{2} = {3-5,7-9}{},
  hline{1-3,6-7,10} = {-}{},
}
                    & \rotatebox{90}{Clay} & \rotatebox{90}{CLIP} & \rotatebox{90}{CROMA} & \rotatebox{90}{DINOv3} & \rotatebox{90}{DOFA} & \rotatebox{90}{Galileo} & \rotatebox{90}{Prithvi-2.0} & \rotatebox{90}{R50-imgnet} & \rotatebox{90}{R50-rnd} & \rotatebox{90}{TerraFM} & \rotatebox{90}{TerraMind} & \rotatebox{90}{ViT-imgnet} & \rotatebox{90}{ViT-rnd}  \\
                    & Full Fine-tuning (OOD = 0.820×ID -23.8) &        &        &        &        &        &        &        &        &        &        &        &        \\
RESISC45 → UCMerced & -8.9                                    & +19.9  & -1.5   & -3.7   & -8.8   & +13.2  & -6.6   & +30.5  & +9.3   & +25.5  & -4.1   & -5.7   & -1.9   \\
UCMerced → RESISC45 & -8.3                                    & -9.6   & -9.1   & -5.5   & -11.2  & +17.4  & -6.2   & +8.3   & -10.9  & -5.2   & -10.0  & -7.6   & +0.9   \\
Average             & -8.6                                    & +5.1   & -5.3   & -4.6   & -10.0  & +15.3  & -6.4   & +19.4  & -0.8   & +10.1  & -7.1   & -6.7   & -0.5   \\
                    & Frozen Backbone (OOD = 0.647×ID -8.2)   &        &        &        &        &        &        &        &        &        &        &        &        \\
RESISC45 → UCMerced & +4.0                                    & -2.4   & 0.0   & -5.0   & -6.1   & +4.7   & -3.6   & +21.3  & +4.0   & +14.0  & +0.9   & -4.6   & +2.0   \\
UCMerced → RESISC45 & +2.5                                    & -11.2  & +1.4   & -3.2   & -5.5   & +5.1   & -6.5   & +2.4   & -1.0   & -10.0  & +1.4   & -5.9   & +1.0   \\
Average             & +3.2                                    & -6.8   & +0.7   & -4.1   & -5.8   & +4.9   & -5.1   & +11.9  & +1.5   & +2.0   & +1.2   & -5.3   & +1.5   
\end{tblr}
\end{table}

% SOURCE SHIFT TABLE
\begin{table}[htbp]
\centering
\small
\caption{\textbf{Source Shift} Effective Robustness.}
\label{tab:source_shift_rho}
\begin{tblr}{
  colsep=2.5pt,
  cell{2}{2} = {c=13}{c},
  cell{4}{2} = {c=13}{c},
  hlines,
  vline{2} = {3,5}{},
}
                    & \rotatebox{90}{Clay} & \rotatebox{90}{CLIP} & \rotatebox{90}{CROMA} & \rotatebox{90}{DINOv3} & \rotatebox{90}{DOFA} & \rotatebox{90}{Galileo} & \rotatebox{90}{Prithvi-2.0} & \rotatebox{90}{R50-imgnet} & \rotatebox{90}{R50-rnd} & \rotatebox{90}{TerraFM} & \rotatebox{90}{TerraMind} & \rotatebox{90}{ViT-imgnet} & \rotatebox{90}{ViT-rnd}      \\
                    & Full Fine-tuning (OOD = 1.203×ID -47.8) &        &        &        &        &        &        &        &        &        &        &        &        \\
DeepGlobe → DFC2022 & -0.9                                    & +6.9   & -15.1  & +1.0   & +1.9   & +6.0   & -3.1   & +5.2   & -9.1   & +8.0   & 0.0   & +0.4   & -1.0   \\
                    & Frozen Backbone (OOD = 1.023×ID -31.6)  &        &        &        &        &        &        &        &        &        &        &        &        \\
DeepGlobe → DFC2022 & +3.8                                    & +6.7   & -1.4   & -2.6   & +2.5   & -7.3   & -0.6   & -0.5   & +3.0   & +8.5   & -8.1   & -0.5   & -3.6   
\end{tblr}
\end{table}

\newpage

\subsection{Extended Results - Absolute Performance Scores}
We summarize here the absolute model scores for the ID and OOD test sets per task.

\begin{figure*}[htbp]
    \centering
    \hspace*{-3.5cm}
    \includegraphics[width=1.5\textwidth]{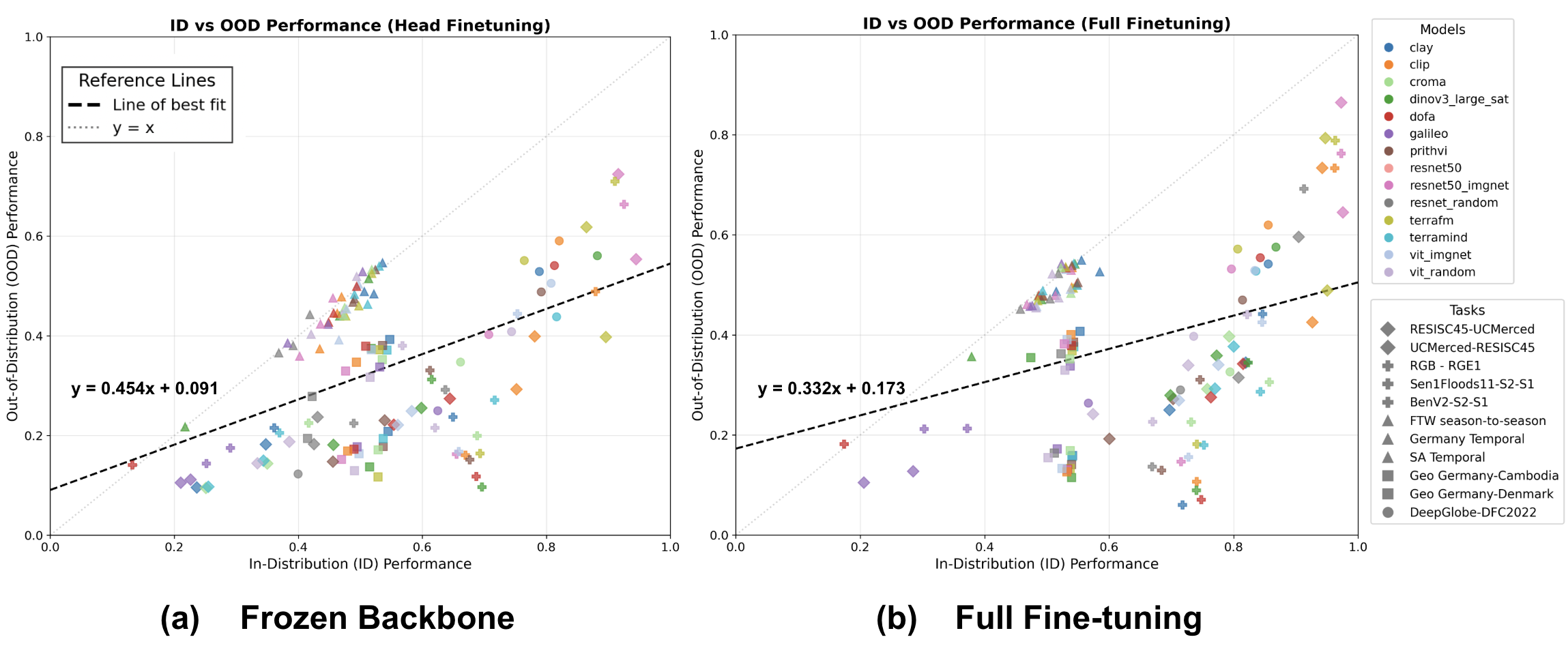}
    \caption{Distribution shift analysis for (a) frozen backbone and (b) full fine-tuning as a function of ID vs OOD absolute performance. Each point represents a model-task combination, with colours denoting model types and shapes denoting tasks. The dashed y=x line represents perfect robustness (no performance drop); the black dashed line represents the line of best fit across all models and tasks. Points indicate mean over 5 random seeds. Shift types are grouped by shape.}
    \label{fig:scatterplots}
\end{figure*}

\begin{figure*}[htbp]
    \centering
    \includegraphics[width=\textwidth]{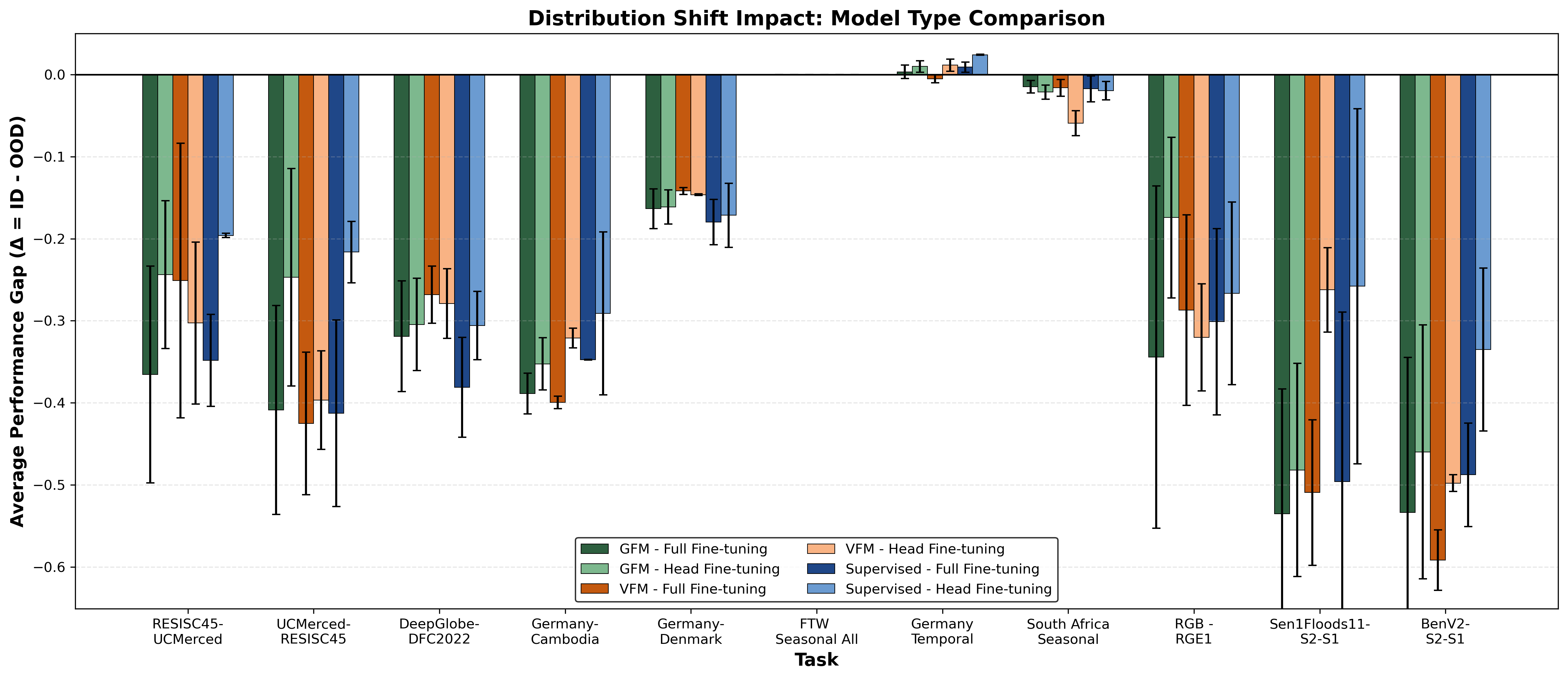}
    \caption{Model-type comparison of absolute difference in OOD-ID test set performance per shift task.}
    \label{fig:barplotdifference}
\end{figure*}

\begin{figure*}[htbp]
    \centering
    \includegraphics[width=\textwidth]{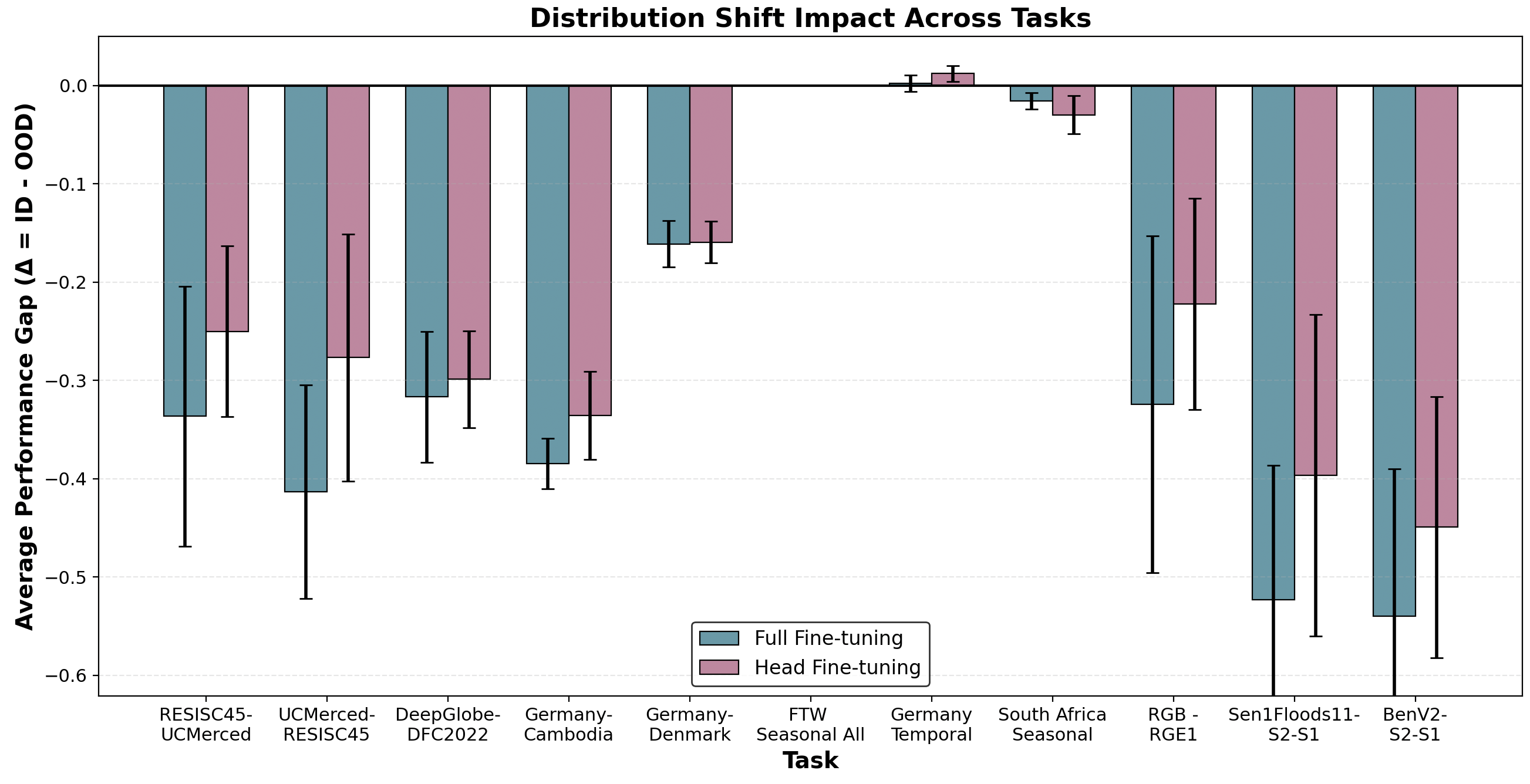}
    \caption{Comparison between full fine-tuning and frozen backbone for absolute performance degradation (OOD-ID) per task.}
    \label{fig:fullfrozen}
\end{figure*}

\begin{figure}[htbp]
  \centering
  % Top Subfigure
  \begin{subfigure}[b]{\textwidth} % Increased width since they are stacked
    \centering
    \includegraphics[width=\linewidth]{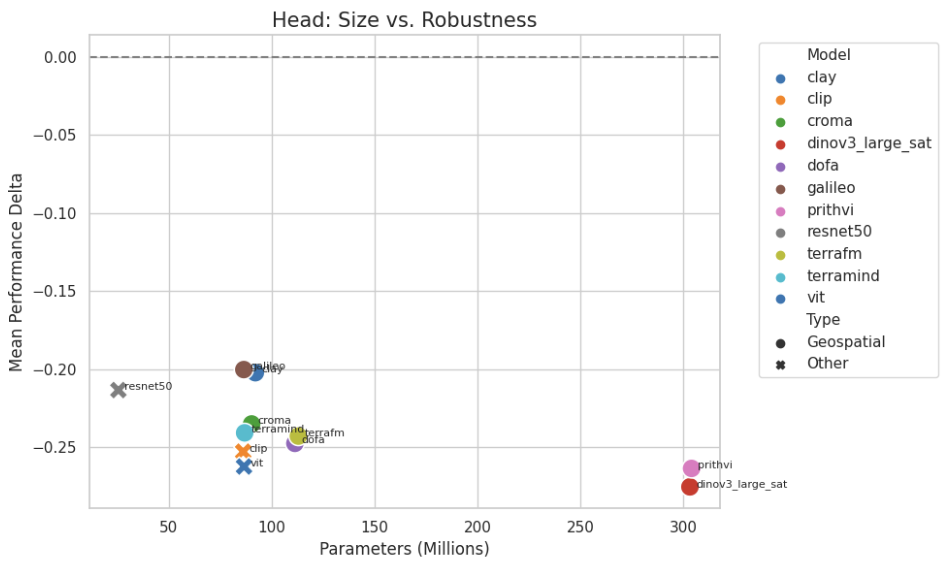}
    \caption{Frozen Backbone}
    \label{fig:param-frozen}
  \end{subfigure}
  
  \vspace{0.5cm} % Adjusts the vertical space between the two images
  
  % Bottom Subfigure
  \begin{subfigure}[b]{\textwidth}
    \centering
    \includegraphics[width=\linewidth]{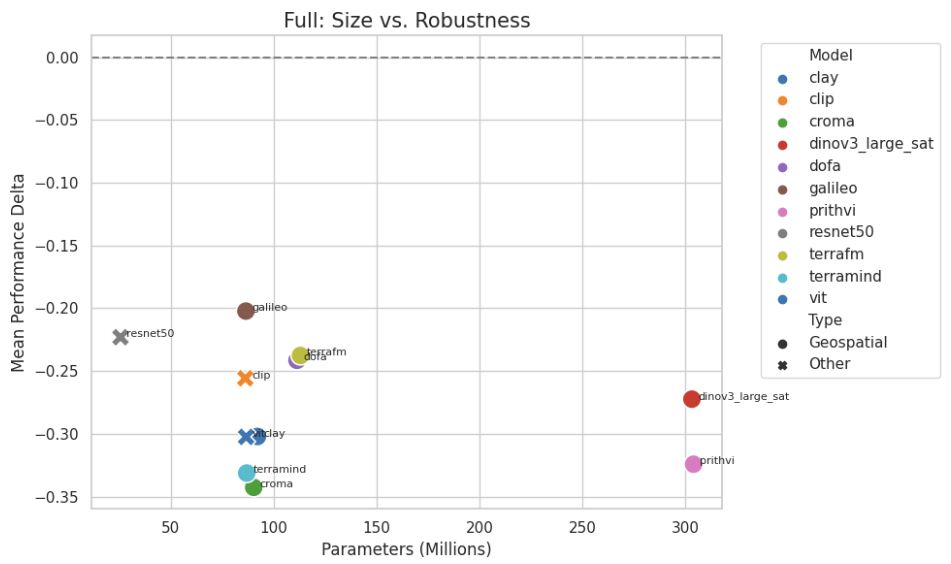}
    \caption{Full Fine-tuning}
    \label{fig:param-full}
  \end{subfigure}

  \caption{\textbf{Model performance as a function of size.} We explore the mean performance delta (OOD-ID score) per model as a function of its backbone parameter size. We show that there is no correlation between model size and mean distributional robustness across our tasks.Model Parameter size vs In Distribution Performance.}
  \label{fig:both_images}
\end{figure}

\begin{table*}[htbp]
\centering
\small
\caption{\textbf{Scale Shift} Results. Stars denote frozen backbone. Scores are using best learning rates per task-model pair, averaged over five random seeds reported with standard deviation across runs. Order of dataset refers to ID-OOD dataset.}
\label{tab:data_shift}
\begin{tabular}{l|ccc|ccc}
\hline
 & \multicolumn{3}{c}{\textbf{RESISC45-UCMerced}} & \multicolumn{3}{c}{\textbf{UCMerced-RESISC45}} \\
\hline
\textbf{Model} & \textbf{ID acc} & \textbf{OOD acc} & $\mathbf{\Delta}$ & \textbf{ID acc} & \textbf{OOD acc} & $\mathbf{\Delta}$ \\
\hline
Clay & 81.94 $\pm$ 0.87 & 34.47 $\pm$ 0.79 & 47.47 $\pm$ 1.17 & 69.71 $\pm$ 1.74 & 25.00 $\pm$ 1.76 & 44.71 $\pm$ 2.48 \\
CROMA & 79.26 $\pm$ 1.73 & 39.67 $\pm$ 0.67 & 39.60 $\pm$ 1.86 & 75.78 $\pm$ 1.34 & 29.19 $\pm$ 1.68 & 46.59 $\pm$ 2.15 \\
DINOv3 & 77.25 $\pm$ 0.64 & 35.85 $\pm$ 0.75 & 41.41 $\pm$ 0.99 & 69.87 $\pm$ 1.46 & 27.94 $\pm$ 1.71 & 41.93 $\pm$ 2.25 \\
Galileo & 28.48 $\pm$ 0.92 & 12.71 $\pm$ 0.48 & 15.77 $\pm$ 1.04 & 20.58 $\pm$ 0.97 & 10.46 $\pm$ 0.53 & 10.12 $\pm$ 1.10 \\
TerraFM & 94.73 $\pm$ 0.35 & 79.32 $\pm$ 2.94 & 15.42 $\pm$ 2.96 & 95.04 $\pm$ 0.82 & 48.88 $\pm$ 3.47 & 46.16 $\pm$ 3.57 \\
Terramind & 80.00 $\pm$ 0.22 & 37.65 $\pm$ 0.60 & 42.35 $\pm$ 0.64 & 76.99 $\pm$ 0.34 & 29.27 $\pm$ 0.70 & 47.73 $\pm$ 0.78 \\
Prithvi-2.0 & 70.26 $\pm$ 2.94 & 27.20 $\pm$ 1.22 & 43.06 $\pm$ 3.18 & 60.05 $\pm$ 4.20 & 19.23 $\pm$ 1.73 & 40.83 $\pm$ 4.54 \\
DOFA & 81.53 $\pm$ 0.50 & 34.19 $\pm$ 0.59 & 47.34 $\pm$ 0.78 & 76.36 $\pm$ 0.71 & 27.54 $\pm$ 0.82 & 48.82 $\pm$ 1.09 \\
ResNet50-random & 90.44 $\pm$ 0.28 & 59.58 $\pm$ 1.87 & 30.86 $\pm$ 1.89 & 80.79 $\pm$ 0.96 & 31.49 $\pm$ 1.63 & 49.30 $\pm$ 1.89 \\
ResNet50-imgnet & 97.28 $\pm$ 0.18 & 86.41 $\pm$ 1.55 & 10.87 $\pm$ 1.57 & 97.57 $\pm$ 0.98 & 64.48 $\pm$ 1.41 & 33.09 $\pm$ 1.71 \\
ViT-random & 72.71 $\pm$ 1.23 & 33.93 $\pm$ 0.69 & 38.77 $\pm$ 1.41 & 57.41 $\pm$ 1.48 & 24.19 $\pm$ 0.58 & 33.22 $\pm$ 1.59 \\
ViT-imgnet & 77.54 $\pm$ 0.29 & 34.00 $\pm$ 0.76 & 43.53 $\pm$ 0.81 & 71.19 $\pm$ 0.57 & 26.90 $\pm$ 0.54 & 44.29 $\pm$ 0.79 \\
CLIP & 94.25 $\pm$ 0.50 & 73.36 $\pm$ 0.52 & 20.89 $\pm$ 0.72 & 92.66 $\pm$ 0.78 & 42.51 $\pm$ 1.33 & 50.15 $\pm$ 1.54 \\
\hline
Clay$^*$ & 34.75 $\pm$ 0.18 & 18.21 $\pm$ 1.4 & 16.54 $\pm$ 1.4 & 23.64 $\pm$ 0.83 & 9.59 $\pm$ 0.31 & 14.05 $\pm$ 0.88 \\
CROMA$^*$ & 34.98 $\pm$ 2.10 & 14.38 $\pm$ 0.75 & 20.60 $\pm$ 2.23 & 25.12 $\pm$ 1.08 & 9.47 $\pm$ 0.73 & 15.65 $\pm$ 1.30 \\
DINOv3$^*$ & 59.86 $\pm$ 0.12 & 25.53 $\pm$ 0.10 & 34.32 $\pm$ 0.16 & 45.65 $\pm$ 0.26 & 18.12 $\pm$ 0.12 & 27.53 $\pm$ 0.29 \\
Galileo$^*$ & 22.64 $\pm$ 1.08 & 11.17 $\pm$ 0.52 & 11.47 $\pm$ 1.20 & 21.06 $\pm$ 1.84 & 10.55 $\pm$ 0.10 & 10.51 $\pm$ 1.84 \\
TerraFM$^*$ & 86.49 $\pm$ 1.82 & 61.78 $\pm$ 6.34 & 24.70 $\pm$ 6.60 & 89.60 $\pm$ 3.81 & 39.72 $\pm$ 4.09 & 49.88 $\pm$ 5.59 \\
Terramind$^*$ & 34.36 $\pm$ 0.31 & 14.94 $\pm$ 0.11 & 19.42 $\pm$ 0.33 & 25.49 $\pm$ 0.89 & 9.71 $\pm$ 0.45 & 15.78 $\pm$ 0.99 \\
Prithvi-2.0$^*$ & 53.92 $\pm$ 3.38 & 23.03 $\pm$ 1.12 & 30.89 $\pm$ 3.56 & 45.59 $\pm$ 4.65 & 14.77 $\pm$ 1.39 & 30.82 $\pm$ 4.86 \\
DOFA$^*$ & 64.45 $\pm$ 0.17 & 27.39 $\pm$ 0.25 & 37.05 $\pm$ 0.30 & 55.41 $\pm$ 0.99 & 22.16 $\pm$ 0.18 & 33.25 $\pm$ 1.00 \\
ResNet50-random$^*$ & 43.11 $\pm$ 0.47 & 23.70 $\pm$ 0.70 & 19.41 $\pm$ 0.85 & 42.53 $\pm$ 2.09 & 18.25 $\pm$ 0.78 & 24.29 $\pm$ 2.23 \\
ResNet50-imgnet$^*$ & 91.61 $\pm$ 0.21 & 72.39 $\pm$ 1.26 & 19.22 $\pm$ 1.28 & 94.46 $\pm$ 0.77 & 55.33 $\pm$ 0.29 & 39.13 $\pm$ 0.82 \\
ViT-random$^*$ & 38.54 $\pm$ 0.82 & 18.76 $\pm$ 0.52 & 19.79 $\pm$ 0.97 & 33.40 $\pm$ 1.43 & 14.43 $\pm$ 0.36 & 18.98 $\pm$ 1.47 \\
ViT-imgnet$^*$ & 58.29 $\pm$ 0.23 & 24.89 $\pm$ 0.29 & 33.40 $\pm$ 0.37 & 56.04 $\pm$ 0.66 & 22.09 $\pm$ 0.36 & 33.95 $\pm$ 0.76 \\
CLIP$^*$ & 78.11 $\pm$ 0.14 & 39.90 $\pm$ 0.39 & 38.21 $\pm$ 0.42 & 75.20 $\pm$ 0.41 & 29.28 $\pm$ 0.30 & 45.91 $\pm$ 0.51 \\
\hline
\end{tabular}
\end{table*}
% data shift table with deepglobe
\begin{table*}[htbp]
\centering
\small
\caption{\textbf{Data Shift} Results for DeepGlobe-DFC2022. Stars denote frozen backbone. Scores are using best learning rates per task-model pair, averaged over five random seeds reported with standard deviation across runs.}
\label{tab:data_shift_deepglobe}
\begin{tabular}{l|ccc}
\hline
 & \multicolumn{3}{c}{\textbf{DeepGlobe-DFC2022}} \\
\hline
\textbf{Model} & \textbf{ID mIoU} & \textbf{OOD mIoU} & $\mathbf{\Delta}$ \\
\hline
Clay & 85.54 $\pm$ 0.25 & 54.19 $\pm$ 1.98 & 31.35 $\pm$ 2.00 \\
CROMA & 79.40 $\pm$ 1.83 & 32.63 $\pm$ 3.32 & 46.77 $\pm$ 3.79 \\
DINOv3 & 86.80 $\pm$ 0.19 & 57.57 $\pm$ 6.34 & 29.22 $\pm$ 6.35 \\
Galileo & 56.63 $\pm$ 2.24 & 26.36 $\pm$ 1.55 & 30.27 $\pm$ 2.72 \\
TerraFM & 80.65 $\pm$ 1.02 & 57.20 $\pm$ 4.12 & 23.45 $\pm$ 4.24 \\
Terramind & 83.58 $\pm$ 0.51 & 52.75 $\pm$ 2.82 & 30.84 $\pm$ 2.87 \\
Prithvi-2.0 & 81.40 $\pm$ 1.82 & 47.00 $\pm$ 11.56 & 34.40 $\pm$ 11.70 \\
DOFA & 84.28 $\pm$ 4.08 & 55.45 $\pm$ 1.59 & 28.83 $\pm$ 4.38 \\
ResNet50-random & 71.43 $\pm$ 1.57 & 29.01 $\pm$ 4.04 & 42.42 $\pm$ 4.34 \\
ResNet50-imgnet & 79.64 $\pm$ 1.55 & 53.19 $\pm$ 3.67 & 26.45 $\pm$ 3.98 \\
ViT-random & 73.56 $\pm$ 1.49 & 39.73 $\pm$ 1.05 & 33.82 $\pm$ 1.82 \\
ViT-imgnet & 83.39 $\pm$ 0.41 & 52.90 $\pm$ 1.89 & 30.49 $\pm$ 1.93 \\
CLIP & 85.54 $\pm$ 0.79 & 61.98 $\pm$ 3.62 & 23.56 $\pm$ 3.70 \\
\hline
Clay$^*$ & 78.86 $\pm$ 0.12 & 52.90 $\pm$ 0.34 & 25.97 $\pm$ 0.36 \\
CROMA$^*$ & 66.15 $\pm$ 2.84 & 34.72 $\pm$ 3.37 & 31.43 $\pm$ 4.41 \\
DINOv3$^*$ & 88.19 $\pm$ 0.23 & 56.06 $\pm$ 1.81 & 32.13 $\pm$ 1.82 \\
Galileo$^*$ & 62.49 $\pm$ 0.22 & 25.00 $\pm$ 0.29 & 37.49 $\pm$ 0.36 \\
TerraFM$^*$ & 76.43 $\pm$ 4.05 & 55.09 $\pm$ 6.35 & 21.33 $\pm$ 7.54 \\
Terramind$^*$ & 81.64 $\pm$ 0.11 & 43.83 $\pm$ 1.17 & 37.81 $\pm$ 1.17 \\
Prithvi-2.0$^*$ & 79.16 $\pm$ 0.68 & 48.83 $\pm$ 5.45 & 30.33 $\pm$ 5.49 \\
DOFA$^*$ & 81.30 $\pm$ 0.16 & 54.11 $\pm$ 0.68 & 27.19 $\pm$ 0.70 \\
ResNet50-random$^*$ & 39.95 $\pm$ 2.73 & 12.31 $\pm$ 1.86 & 27.64 $\pm$ 3.30 \\
ResNet50-imgnet$^*$ & 70.70 $\pm$ 0.37 & 40.25 $\pm$ 0.71 & 30.44 $\pm$ 0.81 \\
ViT-random$^*$ & 74.36 $\pm$ 0.70 & 40.84 $\pm$ 2.41 & 33.52 $\pm$ 2.51 \\
ViT-imgnet$^*$ & 80.76 $\pm$ 0.28 & 50.53 $\pm$ 0.72 & 30.22 $\pm$ 0.77 \\
CLIP$^*$ & 82.05 $\pm$ 0.35 & 59.06 $\pm$ 0.51 & 22.99 $\pm$ 0.62 \\
\hline
\end{tabular}
\end{table*}

% Table 2: Geographic Shift
\begin{table*}[htbp]
\centering
\small
\caption{\textbf{Geographic Shift} Results. Stars denote frozen backbone. Scores are using best learning rates per task-model pair, averaged over five random seeds reported with standard deviation across runs.}
\label{tab:geo_shift}
\begin{tabular}{l|ccc|ccc}
\hline
 & \multicolumn{3}{c}{\textbf{Germany-Cambodia}} & \multicolumn{3}{c}{\textbf{Germany-Denmark}} \\
\hline
\textbf{Model} & \textbf{ID mIoU} & \textbf{OOD mIoU} & $\mathbf{\Delta}$ & \textbf{ID mIoU} & \textbf{OOD mIoU} & $\mathbf{\Delta}$ \\
\hline
Clay & 54.12 $\pm$ 0.18 & 15.88 $\pm$ 0.78 & 38.24 $\pm$ 0.80 & 55.33 $\pm$ 0.13 & 40.71 $\pm$ 0.64 & 14.62 $\pm$ 0.66 \\
CROMA & 53.72 $\pm$ 0.24 & 16.86 $\pm$ 0.97 & 36.86 $\pm$ 1.00 & 53.67 $\pm$ 0.17 & 35.21 $\pm$ 0.72 & 18.46 $\pm$ 0.74 \\
DINOv3 & 53.99 $\pm$ 0.58 & 11.47 $\pm$ 2.75 & 42.52 $\pm$ 2.81 & 47.42 $\pm$ 14.59 & 35.46 $\pm$ 12.47 & 11.96 $\pm$ 19.19 \\
Galileo & 51.63 $\pm$ 0.82 & 17.21 $\pm$ 0.97 & 34.42 $\pm$ 1.27 & 53.75 $\pm$ 0.43 & 33.77 $\pm$ 1.74 & 19.97 $\pm$ 1.79 \\
TerraFM & 53.75 $\pm$ 0.14 & 13.28 $\pm$ 1.99 & 40.47 $\pm$ 1.99 & 54.04 $\pm$ 0.09 & 36.81 $\pm$ 1.23 & 17.23 $\pm$ 1.23 \\
Terramind & 54.00 $\pm$ 0.19 & 15.67 $\pm$ 0.86 & 38.34 $\pm$ 0.88 & 54.34 $\pm$ 0.24 & 37.64 $\pm$ 0.24 & 16.69 $\pm$ 0.34 \\
Prithvi-2.0 & 53.97 $\pm$ 0.20 & 14.07 $\pm$ 1.79 & 39.90 $\pm$ 1.80 & 54.30 $\pm$ 0.18 & 38.53 $\pm$ 0.59 & 15.77 $\pm$ 0.62 \\
DOFA & 53.30 $\pm$ 0.25 & 13.18 $\pm$ 1.28 & 40.12 $\pm$ 1.31 & 53.93 $\pm$ 0.14 & 37.88 $\pm$ 0.67 & 16.04 $\pm$ 0.68 \\
ResNet50-random & 51.18 $\pm$ 0.65 & 16.39 $\pm$ 1.24 & 34.78 $\pm$ 1.40 & 52.27 $\pm$ 0.32 & 36.23 $\pm$ 0.37 & 16.04 $\pm$ 0.49 \\
ResNet50-imgnet & 53.22 $\pm$ 0.83 & 13.02 $\pm$ 1.56 & 40.21 $\pm$ 1.77 & 52.81 $\pm$ 0.18 & 38.20 $\pm$ 0.44 & 14.62 $\pm$ 0.47 \\
ViT-random & 50.19 $\pm$ 1.09 & 15.48 $\pm$ 1.37 & 34.71 $\pm$ 1.75 & 52.90 $\pm$ 0.62 & 32.97 $\pm$ 1.38 & 19.93 $\pm$ 1.51 \\
ViT-imgnet & 52.42 $\pm$ 0.80 & 13.33 $\pm$ 1.22 & 39.09 $\pm$ 1.46 & 53.23 $\pm$ 0.34 & 39.07 $\pm$ 1.14 & 14.17 $\pm$ 1.19 \\
CLIP & 53.21 $\pm$ 0.58 & 12.65 $\pm$ 3.89 & 40.56 $\pm$ 3.93 & 53.85 $\pm$ 0.20 & 40.06 $\pm$ 0.96 & 13.80 $\pm$ 0.98 \\
\hline
Clay$^*$ & 54.47 $\pm$ 0.02 & 20.81 $\pm$ 0.05 & 33.65 $\pm$ 0.06 & 54.75 $\pm$ 0.08 & 39.23 $\pm$ 0.11 & 15.52 $\pm$ 0.14 \\
CROMA$^*$ & 52.89 $\pm$ 0.28 & 17.12 $\pm$ 0.40 & 35.76 $\pm$ 0.49 & 53.53 $\pm$ 0.28 & 35.22 $\pm$ 0.93 & 18.31 $\pm$ 0.97 \\
DINOv3$^*$ & 51.48 $\pm$ 0.88 & 13.70 $\pm$ 3.48 & 37.78 $\pm$ 3.59 & 51.77 $\pm$ 1.18 & 37.43 $\pm$ 1.35 & 14.35 $\pm$ 1.79 \\
Galileo$^*$ & 49.47 $\pm$ 1.05 & 17.73 $\pm$ 1.28 & 31.74 $\pm$ 1.66 & 53.12 $\pm$ 0.50 & 33.75 $\pm$ 0.43 & 19.38 $\pm$ 0.66 \\
TerraFM$^*$ & 52.86 $\pm$ 0.33 & 11.66 $\pm$ 1.95 & 41.20 $\pm$ 1.98 & 53.00 $\pm$ 0.25 & 37.18 $\pm$ 0.88 & 15.82 $\pm$ 0.91 \\
Terramind$^*$ & 53.69 $\pm$ 0.03 & 19.38 $\pm$ 0.17 & 34.31 $\pm$ 0.18 & 54.26 $\pm$ 0.03 & 37.12 $\pm$ 0.17 & 17.14 $\pm$ 0.17 \\
Prithvi-2.0$^*$ & 53.70 $\pm$ 0.22 & 17.79 $\pm$ 1.75 & 35.90 $\pm$ 1.76 & 53.57 $\pm$ 0.19 & 38.04 $\pm$ 0.30 & 15.54 $\pm$ 0.36 \\
DOFA$^*$ & 48.99 $\pm$ 0.32 & 17.27 $\pm$ 2.23 & 31.71 $\pm$ 2.26 & 50.87 $\pm$ 0.05 & 37.92 $\pm$ 0.18 & 12.95 $\pm$ 0.18 \\
ResNet50-random$^*$ & 41.49 $\pm$ 1.72 & 19.40 $\pm$ 0.97 & 22.09 $\pm$ 1.97 & 42.23 $\pm$ 3.06 & 27.84 $\pm$ 2.23 & 14.39 $\pm$ 3.78 \\
ResNet50-imgnet$^*$ & 46.96 $\pm$ 0.90 & 15.17 $\pm$ 2.63 & 31.79 $\pm$ 2.78 & 47.64 $\pm$ 0.45 & 32.91 $\pm$ 0.47 & 14.74 $\pm$ 0.65 \\
ViT-random$^*$ & 49.05 $\pm$ 1.72 & 12.94 $\pm$ 1.56 & 36.11 $\pm$ 2.32 & 51.59 $\pm$ 0.82 & 31.69 $\pm$ 0.50 & 19.90 $\pm$ 0.96 \\
ViT-imgnet$^*$ & 49.78 $\pm$ 0.51 & 16.36 $\pm$ 3.15 & 33.42 $\pm$ 3.19 & 51.75 $\pm$ 0.49 & 37.21 $\pm$ 0.78 & 14.54 $\pm$ 0.92 \\
CLIP$^*$ & 47.96 $\pm$ 2.01 & 16.85 $\pm$ 3.78 & 31.11 $\pm$ 4.28 & 49.35 $\pm$ 1.02 & 34.67 $\pm$ 1.42 & 14.68 $\pm$ 1.75 \\
\hline
\end{tabular}
\end{table*}

% Table 3: Temporal Shift
\begin{table*}[htbp]
\centering
\small
\caption{\textbf{Temporal Shift} Results. Stars denote frozen backbone. Scores are using best learning rates per task-model pair, averaged over five random seeds reported with standard deviation across runs.}
\label{tab:temporal_shift}
\begin{tabular}{l|ccc|ccc}
\hline
 & \multicolumn{3}{c}{\textbf{South Africa Seasonal}} & \multicolumn{3}{c}{\textbf{Germany Year-to-Year}} \\
\hline
\textbf{Model} & \textbf{ID mIoU} & \textbf{OOD mIoU} & $\mathbf{\Delta}$ & \textbf{ID mIoU} & \textbf{OOD mIoU} & $\mathbf{\Delta}$ \\
\hline
Clay & 51.48 $\pm$ 0.49 & 48.68 $\pm$ 1.24 & 2.80 $\pm$ 1.33 & 55.56 $\pm$ 0.21 & 54.91 $\pm$ 0.36 & 0.65 $\pm$ 0.42 \\
CROMA & 48.52 $\pm$ 0.50 & 47.67 $\pm$ 0.31 & 0.85 $\pm$ 0.59 & 52.21 $\pm$ 0.30 & 53.41 $\pm$ 0.16 & 1.20 $\pm$ 0.34 \\
DINOv3 & 49.14 $\pm$ 0.66 & 47.03 $\pm$ 2.56 & 2.11 $\pm$ 2.64 & 54.51 $\pm$ 0.35 & 54.15 $\pm$ 0.16 & 0.36 $\pm$ 0.38 \\
Galileo & 47.30 $\pm$ 3.17 & 45.71 $\pm$ 3.73 & 1.59 $\pm$ 4.89 & 52.35 $\pm$ 0.43 & 54.18 $\pm$ 0.27 & 1.83 $\pm$ 0.51 \\
TerraFM & 48.53 $\pm$ 0.45 & 46.86 $\pm$ 1.00 & 1.67 $\pm$ 1.10 & 52.99 $\pm$ 0.55 & 53.57 $\pm$ 0.18 & 0.58 $\pm$ 0.58 \\
Terramind & 49.32 $\pm$ 0.35 & 48.82 $\pm$ 0.57 & 0.51 $\pm$ 0.67 & 54.32 $\pm$ 0.17 & 54.23 $\pm$ 0.12 & 0.09 $\pm$ 0.21 \\
Prithvi-2.0 & 49.41 $\pm$ 0.44 & 47.77 $\pm$ 2.02 & 1.64 $\pm$ 2.07 & 54.00 $\pm$ 0.38 & 53.85 $\pm$ 0.16 & 0.15 $\pm$ 0.41 \\
DOFA & 48.72 $\pm$ 0.28 & 47.89 $\pm$ 0.67 & 0.83 $\pm$ 0.72 & 53.06 $\pm$ 0.25 & 53.36 $\pm$ 0.14 & 0.31 $\pm$ 0.29 \\
ResNet50-random & 45.76 $\pm$ 0.72 & 45.13 $\pm$ 0.54 & 0.63 $\pm$ 0.90 & 51.83 $\pm$ 0.38 & 52.28 $\pm$ 0.36 & 0.45 $\pm$ 0.53 \\
ResNet50-imgnet & 46.76 $\pm$ 0.14 & 46.07 $\pm$ 0.36 & 0.69 $\pm$ 0.39 & 53.89 $\pm$ 0.31 & 52.89 $\pm$ 0.15 & 0.99 $\pm$ 0.35 \\
ViT-random & 48.26 $\pm$ 0.67 & 45.39 $\pm$ 1.71 & 2.87 $\pm$ 1.84 & 50.83 $\pm$ 0.21 & 52.18 $\pm$ 0.43 & 1.34 $\pm$ 0.48 \\
ViT-imgnet & 48.35 $\pm$ 0.46 & 45.65 $\pm$ 1.00 & 2.71 $\pm$ 1.10 & 53.09 $\pm$ 1.24 & 53.07 $\pm$ 0.37 & 0.02 $\pm$ 1.29 \\
CLIP & 49.23 $\pm$ 0.40 & 47.77 $\pm$ 1.34 & 1.46 $\pm$ 1.40 & 54.03 $\pm$ 1.03 & 53.44 $\pm$ 0.33 & 0.59 $\pm$ 1.08 \\
\hline
Clay$^*$ & 50.65 $\pm$ 0.23 & 48.88 $\pm$ 0.22 & 1.77 $\pm$ 0.32 & 53.60 $\pm$ 0.11 & 54.62 $\pm$ 0.03 & 1.02 $\pm$ 0.12 \\
CROMA$^*$ & 47.49 $\pm$ 1.12 & 45.62 $\pm$ 1.69 & 1.87 $\pm$ 2.02 & 51.86 $\pm$ 0.36 & 53.29 $\pm$ 0.29 & 1.43 $\pm$ 0.46 \\
DINOv3$^*$ & 46.65 $\pm$ 0.45 & 43.98 $\pm$ 1.56 & 2.67 $\pm$ 1.63 & 51.35 $\pm$ 0.65 & 51.46 $\pm$ 0.50 & 0.12 $\pm$ 0.82 \\
Galileo$^*$ & 44.83 $\pm$ 1.45 & 42.28 $\pm$ 2.42 & 2.54 $\pm$ 2.82 & 50.39 $\pm$ 0.63 & 52.83 $\pm$ 0.56 & 2.44 $\pm$ 0.85 \\
TerraFM$^*$ & 47.68 $\pm$ 0.38 & 43.99 $\pm$ 2.83 & 3.69 $\pm$ 2.85 & 51.83 $\pm$ 0.31 & 52.53 $\pm$ 0.17 & 0.70 $\pm$ 0.36 \\
Terramind$^*$ & 49.05 $\pm$ 0.06 & 48.30 $\pm$ 0.25 & 0.75 $\pm$ 0.26 & 53.05 $\pm$ 0.01 & 53.96 $\pm$ 0.02 & 0.91 $\pm$ 0.03 \\
Prithvi-2.0$^*$ & 49.07 $\pm$ 0.09 & 47.44 $\pm$ 0.94 & 1.63 $\pm$ 0.94 & 52.42 $\pm$ 0.47 & 53.24 $\pm$ 0.39 & 0.81 $\pm$ 0.62 \\
DOFA$^*$ & 44.88 $\pm$ 0.80 & 42.73 $\pm$ 0.85 & 2.15 $\pm$ 1.17 & 49.42 $\pm$ 0.09 & 49.91 $\pm$ 0.09 & 0.50 $\pm$ 0.13 \\
ResNet50-random$^*$ & 39.16 $\pm$ 0.51 & 37.99 $\pm$ 1.18 & 1.17 $\pm$ 1.28 & 41.87 $\pm$ 2.68 & 44.23 $\pm$ 1.84 & 2.36 $\pm$ 3.25 \\
ResNet50-imgnet$^*$ & 40.21 $\pm$ 0.84 & 35.87 $\pm$ 3.19 & 4.34 $\pm$ 3.30 & 45.58 $\pm$ 1.47 & 47.58 $\pm$ 0.29 & 2.00 $\pm$ 1.49 \\
ViT-random$^*$ & 47.08 $\pm$ 0.14 & 44.33 $\pm$ 0.92 & 2.75 $\pm$ 0.93 & 49.42 $\pm$ 0.62 & 51.88 $\pm$ 0.68 & 2.46 $\pm$ 0.92 \\
ViT-imgnet$^*$ & 46.56 $\pm$ 0.54 & 39.18 $\pm$ 1.92 & 7.38 $\pm$ 1.99 & 49.74 $\pm$ 0.82 & 50.36 $\pm$ 0.94 & 0.62 $\pm$ 1.25 \\
CLIP$^*$ & 43.46 $\pm$ 1.68 & 37.38 $\pm$ 2.84 & 6.08 $\pm$ 3.30 & 46.96 $\pm$ 1.11 & 47.79 $\pm$ 0.95 & 0.83 $\pm$ 1.47 \\
\hline
\end{tabular}
\end{table*}

% Table 4: Sensor Shift
\begin{table}[htbp]
\centering
\small
\caption{\textbf{Sen1Floods11 Sensor Shift} Results. Stars denote frozen backbone. Scores are using best learning rates per task-model pair, averaged over five random seeds reported with standard deviation across runs.}
\label{tab:sensor_shift}
\begin{tabular}{l|ccc}
\hline
 & \multicolumn{3}{c}{\textbf{Sen1Floods11 S2 - S1}} \\
\hline
\textbf{Model} & \textbf{ID mIoU} & \textbf{OOD mIoU} & $\mathbf{\Delta}$ \\
\hline
Clay & 83.05 $\pm$ 0.03 & 31.42 $\pm$ 0.17 & 51.64 $\pm$ 0.15 \\
CROMA & 81.07 $\pm$ 0.01 & 15.78 $\pm$ 0.16 & 65.29 $\pm$ 0.15 \\
DINOv3 & 73.10 $\pm$ 0.02 & 45.29 $\pm$ 0.03 & 27.81 $\pm$ 0.03 \\
Galileo & 83.31 $\pm$ 0.01 & 27.49 $\pm$ 0.19 & 55.82 $\pm$ 0.20 \\
TerraFM & 74.83 $\pm$ 0.13 & 22.26 $\pm$ 0.20 & 52.57 $\pm$ 0.29 \\
Terramind & 80.20 $\pm$ 0.01 & 14.22 $\pm$ 0.07 & 65.98 $\pm$ 0.07 \\
Prithvi-2.0 & 80.42 $\pm$ 0.01 & 7.96 $\pm$ 0.00 & 72.46 $\pm$ 0.01 \\
DOFA & 80.88 $\pm$ 0.00 & 44.46 $\pm$ 0.01 & 36.42 $\pm$ 0.01 \\
ResNet50-random & 74.90 $\pm$ 0.02 & 39.92 $\pm$ 0.08 & 34.97 $\pm$ 0.08 \\
ResNet50-imgnet & 77.58 $\pm$ 0.01 & 36.83 $\pm$ 0.15 & 40.74 $\pm$ 0.15 \\
ViT-random & 78.79 $\pm$ 0.02 & 14.58 $\pm$ 0.12 & 64.21 $\pm$ 0.12 \\
ViT-imgnet & 79.81 $\pm$ 0.01 & 23.91 $\pm$ 0.22 & 55.90 $\pm$ 0.22 \\
CLIP & 78.58 $\pm$ 0.03 & 22.33 $\pm$ 0.20 & 56.24 $\pm$ 0.21 \\
\hline
Clay$^*$ & 83.79 $\pm$ 0.00 & 43.71 $\pm$ 0.00 & 40.07 $\pm$ 0.00 \\
CROMA$^*$ & 80.10 $\pm$ 0.01 & 15.02 $\pm$ 0.16 & 65.08 $\pm$ 0.19 \\
DINOv3$^*$ & 64.02 $\pm$ 0.04 & 27.47 $\pm$ 0.17 & 36.56 $\pm$ 0.18 \\
Galileo$^*$ & 81.59 $\pm$ 0.01 & 14.30 $\pm$ 0.16 & 67.30 $\pm$ 0.17 \\
TerraFM$^*$ & 78.07 $\pm$ 0.01 & 43.57 $\pm$ 0.03 & 34.51 $\pm$ 0.03 \\
Terramind$^*$ & 78.95 $\pm$ 0.00 & 23.40 $\pm$ 0.14 & 55.55 $\pm$ 0.14 \\
Prithvi-2.0$^*$ & 77.03 $\pm$ 0.01 & 38.43 $\pm$ 0.17 & 38.61 $\pm$ 0.17 \\
DOFA$^*$ & 63.28 $\pm$ 0.01 & 15.54 $\pm$ 0.06 & 47.74 $\pm$ 0.07 \\
ResNet50-random$^*$ & 54.25 $\pm$ 0.08 & 43.76 $\pm$ 0.00 & 10.49 $\pm$ 0.08 \\
ResNet50-imgnet$^*$ & 60.58 $\pm$ 0.04 & 40.15 $\pm$ 0.05 & 20.43 $\pm$ 0.08 \\
ViT-random$^*$ & 71.93 $\pm$ 0.08 & 30.82 $\pm$ 0.18 & 41.11 $\pm$ 0.23 \\
ViT-imgnet$^*$ & 67.66 $\pm$ 0.04 & 37.55 $\pm$ 0.17 & 30.11 $\pm$ 0.15 \\
CLIP$^*$ & 65.90 $\pm$ 0.08 & 37.66 $\pm$ 0.13 & 28.25 $\pm$ 0.19 \\
\hline
\end{tabular}
\end{table}

\begin{table*}[htbp]
\centering
\small
\caption{\textbf{m-EuroSat Sensor Shift} Results. Stars denote frozen backbone. Scores are using best learning rates per task-model pair, averaged over five random seeds reported with standard deviation across runs.}
\label{tab:sensor_shift}
\begin{tabular}{l|ccc}
\hline
 & \multicolumn{3}{c}{\textbf{RGB - RGE1}} \\
\hline
\textbf{Model} & \textbf{ID acc} & \textbf{OOD acc} & $\mathbf{\Delta}$ \\
\hline
Clay & 84.64 $\pm$ 2.13 & 44.18 $\pm$ 7.52 & 40.46 $\pm$ 7.81 \\
CROMA & 85.72 $\pm$ 0.73 & 30.54 $\pm$ 1.36 & 55.18 $\pm$ 1.55 \\
DINOv3 & 82.32 $\pm$ 1.01 & 34.46 $\pm$ 2.51 & 47.86 $\pm$ 2.70 \\
Galileo & 37.19 $\pm$ 0.57 & 21.27 $\pm$ 0.64 & 15.92 $\pm$ 0.86 \\
TerraFM & 96.33 $\pm$ 0.63 & 78.84 $\pm$ 4.38 & 17.49 $\pm$ 4.42 \\
Terramind & 84.30 $\pm$ 0.96 & 28.63 $\pm$ 2.37 & 55.66 $\pm$ 2.56 \\
Prithvi-2.0 & 74.62 $\pm$ 1.74 & 31.02 $\pm$ 4.05 & 43.59 $\pm$ 4.41 \\
DOFA & 17.43 $\pm$ 6.16 & 18.17 $\pm$ 6.21 & 0.74 $\pm$ 8.75 \\
ResNet50-random & 91.29 $\pm$ 1.62 & 69.20 $\pm$ 3.47 & 22.09 $\pm$ 3.83 \\
ResNet50-imgnet & 97.31 $\pm$ 0.28 & 76.22 $\pm$ 4.95 & 21.08 $\pm$ 4.95 \\
ViT-random & 82.19 $\pm$ 1.31 & 44.04 $\pm$ 2.19 & 38.15 $\pm$ 2.56 \\
ViT-imgnet & 84.56 $\pm$ 0.61 & 42.51 $\pm$ 1.64 & 42.05 $\pm$ 1.74 \\
CLIP & 96.25 $\pm$ 0.23 & 73.29 $\pm$ 1.83 & 22.96 $\pm$ 1.84 \\
\hline
Clay$^*$ & 36.11 $\pm$ 0.82 & 21.55 $\pm$ 0.42 & 14.56 $\pm$ 0.92 \\
CROMA$^*$ & 41.67 $\pm$ 1.69 & 22.49 $\pm$ 1.49 & 19.18 $\pm$ 2.25 \\
DINOv3$^*$ & 61.49 $\pm$ 0.15 & 31.22 $\pm$ 0.42 & 30.26 $\pm$ 0.45 \\
Galileo$^*$ & 29.06 $\pm$ 1.20 & 17.51 $\pm$ 1.20 & 11.55 $\pm$ 1.70 \\
TerraFM$^*$ & 91.08 $\pm$ 1.19 & 70.96 $\pm$ 9.34 & 20.12 $\pm$ 9.41 \\
Terramind$^*$ & 36.97 $\pm$ 0.54 & 20.52 $\pm$ 0.37 & 16.45 $\pm$ 0.65 \\
Prithvi-2.0$^*$ & 61.25 $\pm$ 4.76 & 33.05 $\pm$ 6.99 & 28.19 $\pm$ 8.45 \\
DOFA$^*$ & 13.23 $\pm$ 2.76 & 14.10 $\pm$ 2.07 & 0.86 $\pm$ 3.45 \\
ResNet50-random$^*$ & 63.70 $\pm$ 0.74 & 29.17 $\pm$ 1.36 & 34.54 $\pm$ 1.54 \\
ResNet50-imgnet$^*$ & 92.53 $\pm$ 0.29 & 66.37 $\pm$ 1.19 & 26.16 $\pm$ 1.22 \\
ViT-random$^*$ & 56.81 $\pm$ 0.81 & 38.01 $\pm$ 1.04 & 18.80 $\pm$ 1.32 \\
ViT-imgnet$^*$ & 75.32 $\pm$ 0.45 & 44.42 $\pm$ 0.25 & 30.90 $\pm$ 0.51 \\
CLIP$^*$ & 87.93 $\pm$ 0.25 & 48.88 $\pm$ 0.66 & 39.06 $\pm$ 0.71 \\
\hline
\end{tabular}
\end{table*}

\begin{table*}[htbp]
\centering
\small
\caption{\textbf{BigEarthNetv2 S2-S1 Sensor Shift} Results. Stars denote frozen backbone. Scores are using best learning rates per task-model pair, averaged over five random seeds reported with standard deviation across runs.}
\label{tab:sensor_shift}
\begin{tabular}{l|ccc}
\hline
 & \multicolumn{3}{c}{\textbf{Benv2 S2-S1}} \\
\hline
\textbf{Model} & \textbf{ID F1} & \textbf{OOD F1} & $\mathbf{\Delta}$ \\
\hline
Clay & 71.79$\pm$0.37 & 6.02$\pm$7.19 & -65.77 \\
CROMA & 73.16$\pm$0.32 & 22.59$\pm$6.26 & -50.57 \\
DINOv3 & 74.04$\pm$0.18 & 8.96$\pm$4.05 & -65.08 \\
Galileo & 30.27$\pm$4.66 & 21.20$\pm$9.29 & -9.07 \\
TerraFM & 74.09$\pm$0.36 & 18.16$\pm$4.63 & -55.93 \\
Terramind & 75.22$\pm$0.24 & 17.96$\pm$4.56 & -57.26 \\
Prithvi-2.0 & 68.41$\pm$2.42 & 12.93$\pm$7.97 & -55.47 \\
DOFA & 74.79$\pm$0.47 & 7.09$\pm$3.44 & -67.70 \\
ResNet50-random & 66.90$\pm$1.30 & 13.66$\pm$5.46 & -53.24 \\
ResNet50-imgnet & 71.57$\pm$0.44 & 14.65$\pm$5.41 & -56.92 \\
ViT-random & 66.98$\pm$0.77 & 22.65$\pm$3.43 & -44.33 \\
ViT-imgnet & 72.76$\pm$0.29 & 15.59$\pm$3.70 & -57.17 \\
CLIP & 74.06$\pm$0.36 & 10.65$\pm$3.69 & -63.41 \\
\hline
Clay$^*$ & 64.95$\pm$0.04 & 23.71$\pm$0.02 & -41.24 \\
CROMA$^*$ & 68.86$\pm$0.52 & 19.93$\pm$6.79 & -48.93 \\
DINOv3$^*$ & 69.61$\pm$0.38 & 9.66$\pm$6.04 & -59.95 \\
Galileo$^*$ & 25.17$\pm$1.15 & 14.36$\pm$8.46 & -10.81 \\
TerraFM$^*$ & 69.31$\pm$0.51 & 16.38$\pm$6.77 & -52.93 \\
Terramind$^*$ & 71.68$\pm$0.11 & 27.11$\pm$0.34 & -44.57 \\
Prithvi-2.0$^*$ & 67.64$\pm$0.73 & 15.12$\pm$5.65 & -52.52 \\
DOFA$^*$ & 68.69$\pm$0.06 & 11.78$\pm$0.44 & -56.91 \\
ResNet50-random$^*$ & 48.95$\pm$2.36 & 22.46$\pm$6.72 & -26.49 \\
ResNet50-imgnet$^*$ & 65.53$\pm$0.42 & 16.24$\pm$6.39 & -49.29 \\
ViT-random$^*$ & 62.05$\pm$1.15 & 21.53$\pm$3.49 & -40.52 \\
ViT-imgnet$^*$ & 65.88$\pm$1.68 & 16.78$\pm$6.27 & -49.10 \\
CLIP$^*$ & 66.97$\pm$0.89 & 16.04$\pm$11.35 & -50.93 \\
\hline
\end{tabular}
\end{table*}

\newpage

\begin{table}[htbp]
\centering
\caption{Geographic shift optimal learning rates from learning rate sweep using Germany-Cambodia shift}
\begin{tabular}{l|cc}
\hline
\textbf{Model} & \textbf{Frozen LR} & \textbf{Full Fine-tune LR} \\
\hline
Clay             & 5e-3      & 3e-4              \\
CROMA            & 3e-3      & 6e-5              \\
DINOv3           & 5e-3      & 6e-4              \\
Galileo          & 3e-3      & 6e-3              \\
TerraFM          & 1e-4      & 1e-4              \\
Terramind        & 3e-4      & 3e-4              \\
Dofa             & 5e-4      & 3e-5              \\
Prithvi-2.0          & 3e-3      & 1e-4              \\
Resnet50\_random & 5e-4      & 6e-4              \\
Resnet50\_imgnet & 5e-4      & 3e-3              \\
ViT\_random      & 3-e3      & 6e-5              \\
ViT\_imgnet      & 5e-4      & 6e-5              \\
CLIP             & 5e-4      & 6e-5   \\
\hline
\end{tabular}
\end{table}

\begin{table}[htbp]
\centering
\caption{Scale shift, classification, optimal learning rates from learning rate sweep}
\begin{tabular}{l|cc}
\hline
\textbf{Model} & \textbf{Frozen LR} & \textbf{Full Fine-tune LR} \\
\hline
Clay             & 5e-2      & 3e-5              \\
CROMA            & 5e-2      &  6e-5                \\
DINOv3           & 3e-3      & 1e-4              \\
Galileo          & 1e-3      & 6e-3              \\
TerraFM          & 5e-3      & 3e-5              \\
Terramind        & 5e-2      & 3e-4              \\
DOFA             & 5e-2      & 1e-5              \\
Prithvi-2.0          & 1e-2      & 1e-5              \\
Resnet50\_random & 1e-2      & 3e-3              \\
Resnet50\_imgnet & 1e-3      & 3e-4              \\
ViT\_random      & 3e-2      & 6e-4              \\
ViT\_imgnet      & 5e-2      & 3e-5              \\
CLIP             & 3-e3      & 1e-5            \\
\hline
\end{tabular}
\end{table}

\begin{table}[htbp]
\centering
\caption{Data shift, semantic segmentation, optimal learning rates from learning rate sweep}
\begin{tabular}{l|cc}
\hline
\textbf{Model} & \textbf{Frozen LR} & \textbf{Full Fine-tune LR} \\
\hline
Clay             & 5e-3      & 3e-4              \\
CROMA            & 3e-3      & 6e-5              \\
DINOv3           & 5e-3      & 6e-5              \\
Galileo          & 3e-3      & 1e-3              \\
TerraFM          & 1e-3      & 1e-5              \\
Terramind        & 3e-3      & 3e-5              \\
DOFA             & 1e-3      & 6e-5              \\
Prithvi-2.0          & 5e-3      & 3e-5              \\
Resnet50\_random & 1e-4      & 3e-4              \\
Resnet50\_imgnet & 1e-4      & 1e-3              \\
ViT\_random      & 3-e3      & 6e-5              \\
ViT\_imgnet      & 5e-3      & 3e-5              \\
CLIP             & 5e-3      & 3e-5              \\
\hline
\end{tabular}
\end{table}

\begin{table}[htbp]
\centering
\caption{Sensor shift, optimal learning rates from learning rate sweep using RGB $\rightarrow$ RE1E2 shift}
\begin{tabular}{l|cc}
\hline
\textbf{Model} & \textbf{Frozen LR} & \textbf{Full Fine-tune LR} \\
\hline
Clay             & 5e-2      & 3e-4              \\
CROMA            & 3e-2      & 3e-4              \\
DINOv3           & 1e-2      & 6e-5              \\
Galileo          & 1e-2      & 6e-3              \\
TerraFM          & 1e-3      & 6e-5              \\
Terramind        & 3e-2      & 3e-4              \\
Dofa             & 5e-2      & 6e-5              \\
Prithvi-2.0          & 5e-2      & 3e-5              \\
Resnet50\_random & 1e-2      & 1e-3              \\
Resnet50\_imgnet & 3e-3      & 1e-4              \\
ViT\_random      & 5e-2      & 1e-4              \\
ViT\_imgnet      & 5e-2      & 6e-5              \\
CLIP             & 1e-2      & 1e-5 \\
\hline
\end{tabular}
\end{table}

\begin{table}[htbp]
\centering
\caption{Sensor shift, classification, optimal learning rates from learning rate sweep using S2 $\rightarrow$ S1 BenV2 shift}
\begin{tabular}{l|cc}
\hline
\textbf{Model} & \textbf{Frozen LR} & \textbf{Full Fine-tune LR} \\
\hline
Clay             & 5e-2      & 6e-4           \\
CROMA            & 5e-2      & 1e-4            \\
DINOv3           & 1e-2      & 3e-4              \\
Galileo          & 1e-4      & 1e-5              \\
TerraFM          & 3e-3      & 6e-5           \\
Terramind        & 3e-2      & 6e-5               \\
Dofa             & 1e-2      & 1e-4          \\
Prithvi-2.0          & 3e-2      & 3e-5        \\
Resnet50\_random & 3e-2      & 6e-5        \\
Resnet50\_imgnet & 1e-3      & 1e-3          \\
ViT\_random      & 3e-2      & 1e-4         \\
ViT\_imgnet      & 5e-4      & 1e-3              \\
CLIP             & 3e-3      & 6e-5\\
\hline
\end{tabular}
\end{table}

\begin{table}[htbp]
\centering
\caption{Sensor shift, semantic segmentation, optimal learning rates from learning rate sweep using S2 $\rightarrow$ S1 Sen1Floods11 shift}
\begin{tabular}{l|cc}
\hline
\textbf{Model} & \textbf{Frozen LR} & \textbf{Full Fine-tune LR} \\
\hline
Clay             &   3e-2    &      1e-3    \\
CROMA            &   5e-3    &     6e-5          \\
DINOv3           &   1e-4    &    3e-4           \\
Galileo          &   3e-3    &    6e-3           \\
TerraFM          &   3e-3    &    6e-4           \\
Terramind        &   3e-3    &    1e-4          \\
Dofa             &  3e-4     &    3e-4           \\
Prithvi-2.0          &  5e-3     &    3e-4          \\
Resnet50\_random &  5e-3     &    1e-3          \\
Resnet50\_imgnet &   1e-3    &    3e-3           \\
ViT\_random      &  1e-3     &    1e-4           \\
ViT\_imgnet      &   3e-3    &    1e-3           \\
CLIP             &  5e-3   & 3e-4 \\
\hline
\end{tabular}
\end{table}

\begin{table}[htbp]
\centering
\caption{Temporal shift, South Africa window optimal learning rates from learning rate sweep}
\begin{tabular}{l|cc}
\hline
\textbf{Model} & \textbf{Frozen LR} & \textbf{Full Fine-tune LR} \\
\hline
Clay             & 3e-3      & 6e-4              \\
CROMA            & 3e-3      & 1e-4              \\
DINOv3           & 3e-2      & 1e-3              \\
Galileo          & 5e-3      & 3e-3              \\
TerraFM          & 5e-4      & 1e-4              \\
Terramind        & 5e-4      & 3e-4              \\
Dofa             & 5e-3      & 3e-5              \\
Prithvi-2.0          & 1e-2      & 1e-4              \\
Resnet50\_random & 5e-4      & 6e-4              \\
Resnet50\_imgnet & 1e-4      & 3e-3              \\
ViT\_random      & 5e-3      & 3e-5              \\
ViT\_imgnet      & 3e-3      & 3e-4              \\
CLIP             & 3e-3      & 1e-4     \\
\hline
\end{tabular}
\end{table}

\clearpage
\newpage

\end{document}